\pdfoutput=1
\documentclass[11pt]{article}
\usepackage[]{acl}

\usepackage{times}
\usepackage{latexsym}

\usepackage[T1]{fontenc}
\usepackage[utf8]{inputenc}

\usepackage{microtype}

\usepackage{xspace} 
\usepackage{graphicx}
\usepackage{textcomp}
\usepackage{multirow}
\usepackage{colortbl}
\usepackage{hyperref}   
\usepackage{hhline}
\usepackage{bbding}
\usepackage{makecell}
\usepackage{color}
\usepackage{xcolor}
\usepackage{arydshln} 
\usepackage{booktabs}
\usepackage{amssymb}
\usepackage{vcell}
\usepackage{subcaption}
\usepackage{diagbox}
\usepackage{tabularray}
\usepackage{float}

\newcommand{\attentionextractor}{AttentionExtractor\xspace}
\newcommand{\attentioncoder}{AttentionCoder\xspace}
\newcommand{\attention}{\textit{Attention}\xspace}
\newcommand{\multinlh}{MultiNL-H\xspace}

\definecolor{greencolor}{RGB}{55,126,34}
\definecolor{yellowcolor}{RGB}{191,144,0}
\definecolor{darkgreen}{RGB}{0, 128, 0}

\title{Improving Natural Language Capability of Code Large Language Model}

\author{Wei Li$^{1,2}$\thanks{~~Equal contribution}, Daoguang Zan$^{1,2*\mathcal{y}}$, Bei Guan$^{2,4}$\thanks{~~Corresponding authors}, Ailun Yu$^{3}$, Xiaolin Chen$^{1,2}$, Yongji Wang$^{2,4}$\\
  $^1$Cooperative Innovation Center, Institute of Software, Chinese Academy of Sciences \\
  $^2$University of Chinese Academy of Sciences; 
  $^3$Peking University; \\
  $^4$Integrative Innovation Center, Institute of Software, Chinese Academy of Sciences \\
  \texttt{\{liwei@,daoguang@,guanbei@,chenxiaolin2019@,ywang@itechs.\}iscas.ac.cn}\\
}

\begin{document}
\maketitle
\begin{abstract}
Code large language models (Code LLMs) have demonstrated remarkable performance in code generation.
Nonetheless, most existing works focus on boosting code LLMs from the perspective of programming capabilities, while their natural language capabilities receive less attention.
To fill this gap, we thus propose a novel framework, comprising two modules:
\attentionextractor, which is responsible for extracting key phrases from the user's natural language requirements, 
and \attentioncoder, which leverages these extracted phrases to generate target code to solve the requirement. 
This framework pioneers an innovative idea by seamlessly integrating code LLMs with traditional natural language processing tools.
To validate the effectiveness of the framework, we craft a new code generation benchmark, called \multinlh, covering five natural languages.
Extensive experimental results demonstrate the effectiveness of our proposed framework.
\end{abstract}

\section{Introduction}
\label{Introduction}

Code Large Language Models (code LLMs) have demonstrated remarkable proficiency in code generation tasks~\cite{nl2code,zhang2023unifying,zheng2023survey}, which can transform users' natural language requirements into target code using various programming languages. 
Numerous methods have emerged to enhance the performance of code LLMs, achieving significant advancements by constructing high-quality training data~\cite{wizardcoder,pangucoder2,phind} or well-designed prompt~\cite{cot,shinn2023reflexion}.
However, previous works predominantly put efforts into enhancing code LLMs from the perspective of programming capabilities, rarely directing attention to the natural language understanding abilities.
The precondition for generating correct code is that the model can accurately comprehend the programming task, which commonly involves natural language understanding.
The practice of training code LLMs on a corpus comprising both code and natural language training instances, which leads to a powerful model CodeLlama~\cite{codellama}, implies the important role of natural language understanding abilities in code LLMs.
Therefore, this paper will delve into how to help code LLMs comprehend natural language context more accurately, thereby boosting the code generation capabilities of code LLMs.

\begin{figure}[t]
    \centering
    \includegraphics[width=0.75\linewidth]{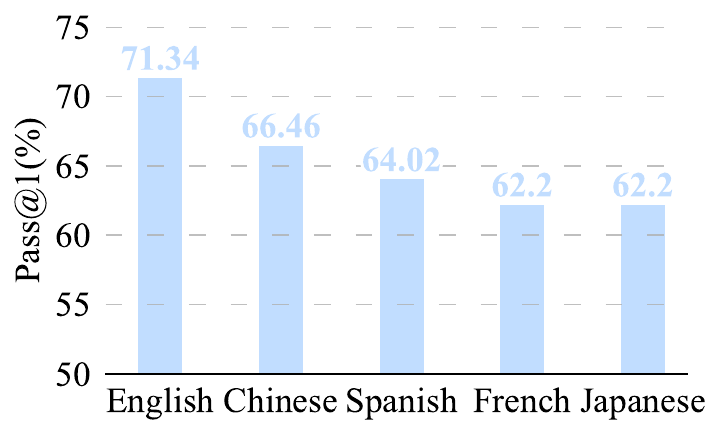}
    \caption{Pass$@1$ of OpenAI's GPT-3.5-turbo evaluated on HumanEval across multiple natural languages.}
    \label{fig:pre_experiment}
\end{figure}

After conducting preliminary experiments on HumanEval~\cite{codex} across multiple natural languages, we find that code LLMs exhibit notable variations between different natural languages in code generation performance.
Figure~\ref{fig:pre_experiment} demonstrates a significant drop in code generation performance for other languages compared to English.
This highlights the need for more exploration of code generation in various natural languages.
To meet this challenge, we propose a framework to guide code LLMs to focus on critical information embedded in the given natural language context, helping code LLMs to comprehend the task well.
From the perspective of cognitive science, the cognitive load theory~\cite{sweller1988cognitive,sweller2011cognitive,kalyuga2011cognitive} shows that humans inherently prioritize key information during handling new scenarios.
Inspired by this theory, our framework highlights important words, phrases, and sentences in the task description for code LLMs to facilitate the models in more easily capturing the key information in given programming requirements.
This framework comprises two modules: \attentionextractor, which identifies pivotal segments of natural-language task description as \textit{Attention} (e.g., words, phrases, or sentences), and \attentioncoder, which utilizes the extracted \textit{Attention} to generate the target code by harnessing the capabilities of off-the-shelf pre-trained models.

In order to comprehensively evaluate our approach, we craft a multi-lingual benchmark, namely \multinlh, adapted from the HumanEval~\cite{codex} which describes code tasks only in English.
\multinlh comprises code generation tasks described with multiple natural languages including English, Chinese, French, Japanese, and Spanish.
On \multinlh, we conduct extensive experiments to validate whether our framework can boost code LLMs in task understanding across different natural languages.
The results show that our approach exhibits obvious advantages over its baselines. 
For example, by equipping code LLMs with \attentionextractor and \attentioncoder,  we boost OpenAI's GPT-3.5-turbo in code generation for Chinese tasks by more than an absolute improvement of $10\%$ on pass$@1$. 

In summary, the contributions of this work are threefold:

\begin{itemize}
    \item We propose a simple yet effective framework, composed of the \attentionextractor and \attentioncoder modules, to boost the code generation capabilities of code LLMs across multiple natural languages. This framework can inspire researchers to solve tasks by integrating code LLMs with traditional natural language processing (NLP) analysis tools.
    \item We craft a new benchmark, namely \multinlh, which extends the HumanEval (English) benchmark into four natural languages. It can evaluate the code generation capabilities of code LLMs from the perspective of natural languages.
    \item We perform extensive experiments to prove the effectiveness and superiority of our approach, as well as to derive some valuable insights. 
    Our work including all data and code can be seen on \url{https://github.com/NL2Code/AttentionCoder}.
\end{itemize}
\section{Framework}
\label{sec:Framework}

We would define the code generation task.
Given a \textit{context} that contains the programming problem, this task aims to output the \textit{target code} to solve it.
This paper leverages code LLM (defined as $\mathcal{M}$) to address this task.
Formally, let \(\textbf{x}\) represent the context.
Given $\textbf{x}$, the code generation task can be formulated as $\textbf{y}=\mathcal{M}(\textbf{x})$, where \(\textbf{y}\) represents the target code.

\begin{figure}
    \centering
    \includegraphics[width=1\linewidth]{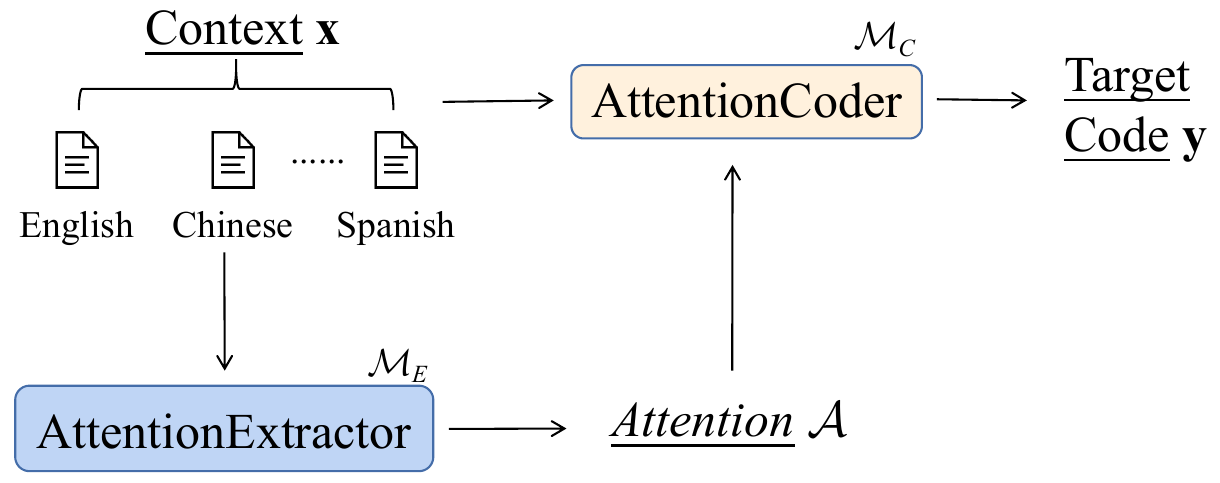}
    \caption{The overview of our proposed framework.}
    \label{fig: framework}
\end{figure}

In this paper, we propose a framework to solve this task, and it includes two modules: \attentionextractor ($\mathcal{M}_E$) and \attentioncoder ($\mathcal{M}_C$), as depicted in Figure~\ref{fig: framework}.
\attentionextractor first identifies and extracts key information $\mathcal{A}$ (e.g., words, phrases, or sentences) from the programming problem in the given context $\textbf{x}$.
Subsequently, \attentioncoder focuses on generating the target code $\textbf{y}$ by leveraging $\mathcal{A}$ and $\textbf{x}$.
Therefore, the code generation task can be also reformulated as $\textbf{y}=\mathcal{M}(\textbf{x},\mathcal{A})$.
Specifically, the entire process of our framework can be formalized as \(\mathcal{A} = \mathcal{M}_E(\textbf{x})\) and \(\textbf{y} = \mathcal{M}_C(\textbf{x}, \mathcal{A})\).
Note that \(\mathcal{A}\) is called ``\textit{Attention}'' in our experiments.

\begin{figure*}[ht]
    \centering
    \includegraphics[width=0.8\linewidth]{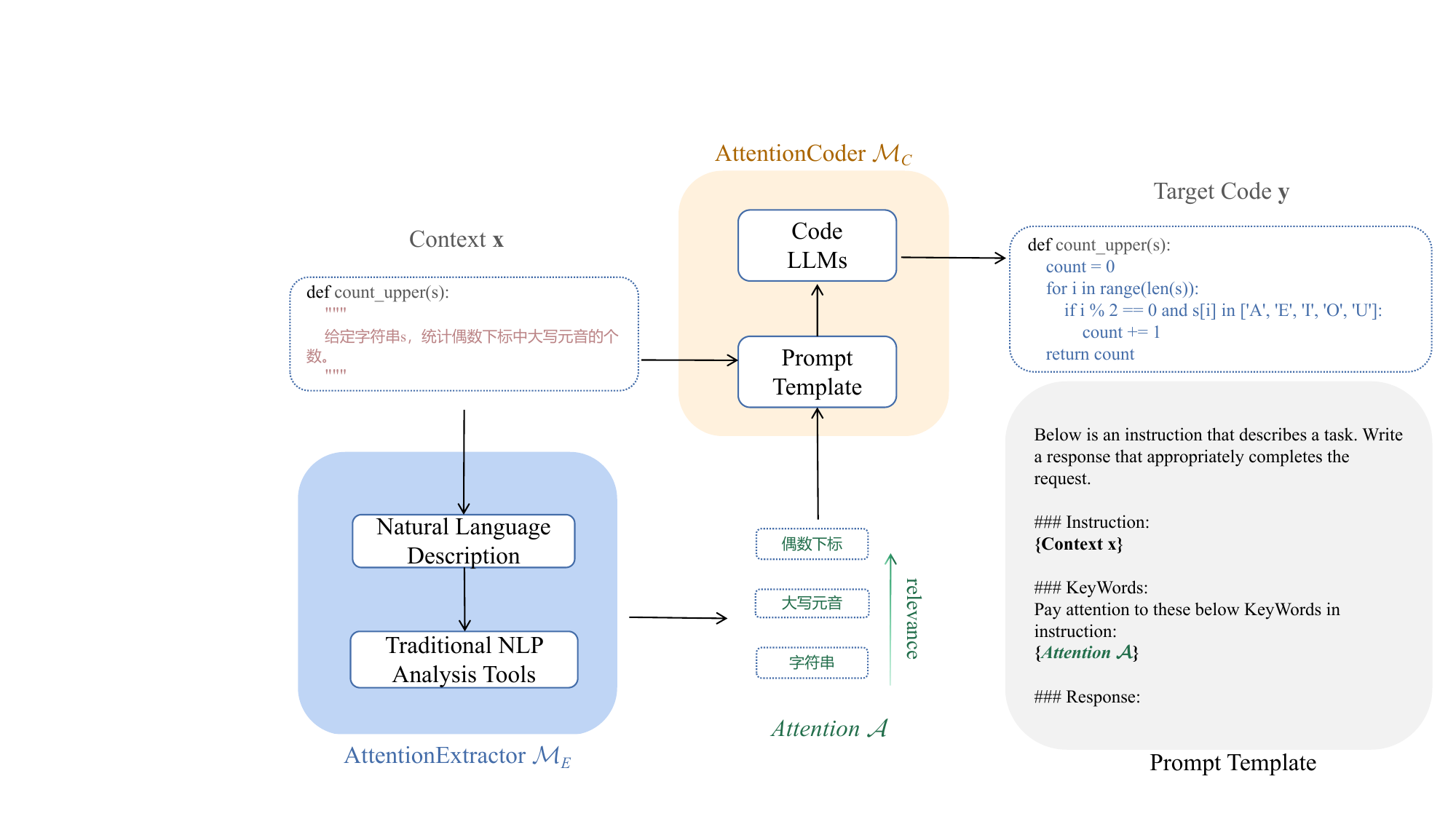}
    \caption{The implementation details of our framework: \attentionextractor and \attentioncoder.}
    \label{fig:complete_details}
\end{figure*}

\section{Methodology}
\label{Methodology}

This section will implement our proposed framework to enhance the multiple natural language capabilities of code LLMs, including \attentionextractor and \attentioncoder.

\subsection{\attentionextractor}

The goal of this module is to extract key information (alias ``\textit{Attention}'') from natural language description.
In the code generation scenario, as shown in Figure~\ref{fig:complete_details}, \attentionextractor first extracts the code comment in the given \textit{context} $\textbf{x}$ using regular expression.
Subsequently, \attentionextractor tokenizes the extracted code comment and performs part-of-speech(POS) tagging using \texttt{spaCy} model\footnote{\url{https://spacy.io}}.
Based on these outcomes, \attentionextractor employs the key phrase extraction algorithm such as TextRank~\cite{mihalcea2004textrank} implemented by PKE package\footnote{\url{https://github.com/boudinfl/pke.git}} to extract multiple phrases\footnote{The POS patterns designed for PKE package is ``\texttt{NP: <NOUN|PROPN|ADJ>*<NOUN|PROPN>; VP: <VERB><NP>}`` defined by using regular expression grammar and POS tag.} as \textit{Attention} and ranks them according to their relevance in the text network.
For instance, \attention extracted from the code comment are three Chinese phases in Figure~\ref{fig:complete_details}. 
These extracted \attention will aid \attentioncoder in generating more accurate code.
Besides using the automatic extraction method above, we also implement \attentionextractor by inviting five participants to manually extract key phrases for each problem.

\subsection{\attentioncoder}

\attentionextractor has successfully extracted \attention for a programming problem, and then \attentioncoder aims to leverage these \attention to generate target code.
We make use of the most straightforward way for \attentioncoder: considering both these extracted \attention $\mathcal{A}$ and the context $\textbf{x}$ when prompting models.
Formally, the \attentioncoder can be written as $\textbf{y}=\mathcal{M}_C(\mathcal{T}(\textbf{x},\mathcal{A}))$, where $\mathcal{T}(,)$ denotes the prompt template that can concatenate  $\textbf{x}$ and $\mathcal{A}$.
The detailed templates can be found in Figure~\ref{fig:complete_details}.
In our experiments, the off-the-shelf code generation models, such as OpenAI's GPT-4~\cite{gpt-4}, GPT-3.5-turbo~\cite{instructgpt}, WizardCoder~\cite{wizardcoder}, and CodeLlama~\cite{codellama}, can be applied directly to implement our \attentioncoder.
\section{Benchmark Construction}
\label{benchmark}

To comprehensively evaluate the efficacy of our framework, we carefully adapt HumanEval~\cite{codex} from English to four other natural languages, including Chinese, French, Spanish, and Japanese, and craft a new benchmark called \multinlh (\textbf{Multi}ple \textbf{N}atural \textbf{L}anguage \textbf{H}umanEval).
Specifically, we invite four native speakers to translate the code comment from English into four other languages.
After careful checking, we ensure the coherence and fluency of the translated content, as well as its consistency with the original content.
Furthermore, we also invite four Python developers who are native speakers of the corresponding natural language to solve the programming problems in translated HumanEval.
Upon their feedback, we further polish and refine this benchmark.
Overall, our meticulous translation process ensures a high-quality benchmark for multi-natural-language-oriented code generation.
We hope that this benchmark will contribute to the rapid advancement of the field.

\section{Experiments}
In this section, we evaluate our framework by answering three research questions (RQs):
\begin{itemize}
    \item \textbf{RQ1 (Effectiveness Validation)}: How is the performance of our framework on multiple-natural-languages-oriented code generation?
    \item \textbf{RQ2 (Ablation Study)}: What factors influence the effectiveness of our framework?
    \item \textbf{RQ3 (Transferability Evaluation)}: Whether our framework is transferable to other tasks involving natural-language context, by reusing \attentionextractor?
\end{itemize}

We answer RQ1 by applying our framework to various models and comparing their performance before and after the application of \attention.
To address RQ2, we conduct extensive ablation studies to examine the impact of various factors on our framework's effectiveness.
We answer RQ3 by extending our framework to other downstream tasks, including multiple-programming-languages-oriented code generation, code translation, and mathematical reasoning.
Specifically, we reuse the task-agnostic \attentionextractor to adapt to different downstream tasks.
\subsection{Experimental Setup}

\paragraph{Benchmark}
We evaluate our framework on our newly proposed \multinlh as described in Section~\ref{benchmark}.
For each evaluation instance, \multinlh provides five different language versions of the code prompt (English, Chinese, French, Spanish, and Japanese) as well as the corresponding canonical solution and test cases.

\paragraph{Baselines}
We choose several existing code LLMs as our base models, including close-source (OpenAI's GPT-4-0613 and GPT-3.5-turbo-0613) and open-source ones (WizardCoder $1$B, $3$B, $7$B, and $15$B; CodeLlama $7$B and $13$B).
We directly prompt these models with different natural languages provided by \multinlh, obtaining their performance without \attention as baseline results.
Additionally, we compare other competitive prompt engineering-based approaches:
\begin{itemize}
    \item \textbf{Chain of Thought (CoT)}~\cite{cot,cot_code}: CoT improves the capability of code LLMs by prompting them to think step by step.
    \item \textbf{Reflexion}~\cite{shinn2023reflexion}: Reflexion augments code LLMs by fostering reflective analysis in prompts, thereby enhancing task comprehension and code generation precision.
\end{itemize}

\paragraph{Evaluation Metrics}
Following the approach outlined by Codex~\cite{codex}, we employ the pass$@1$ metric to assess the performance of all models.
Each model is tasked with generating a single code solution to every programming problem using greedy decoding. The generated solution is then tested against given test cases.
A programming problem is deemed to be successfully solved only if the generated code solution passes all the test cases.
Under this, pass$@1$ is defined as $\frac{|P_{s}|}{|P|}$, where $|P|$ signifies the total count of programming problems, and $|P_{s}|$ indicates the number of challenges successfully resolved.
Thus, the pass$@1$ metric also can be served as the accuracy in our evaluation.

\paragraph{Implementation Details}
In the implementation of \attentionextractor, we treat all extracted noun and verb phrases as \attention.
We implement \attentioncoder using existing models, employ the greedy decoding strategy, and set \texttt{max\_new\_tokens} to $1024$. Furthermore, post-processing is applied to extract relevant code from chat-style responses to ensure a fair evaluation.
For details on reproducing the CoT and Reflexion, please see Appendix~\ref{apx:other_method_template}.

\begin{table*}[!t]
\small
\centering
\setlength{\extrarowheight}{0pt}
\addtolength{\extrarowheight}{\aboverulesep}
\addtolength{\extrarowheight}{\belowrulesep}
\setlength{\aboverulesep}{0pt}
\setlength{\belowrulesep}{0pt}
\begin{tabular}{c|l|lllll} 
\toprule
\multicolumn{1}{c|}{\textbf{Model}}                    & \textbf{Settings}                                     & \textbf{English}                                                  & \textbf{Chinese}                                                  & \textbf{French}                                                   & \textbf{Spanish}                                                  & \textbf{Japanese}                                                  \\ 
\hline
\multirow{3}{*}{WizardCoder 1B}               & {\cellcolor[rgb]{0.757,0.867,1}}No Attention & {\cellcolor[rgb]{0.757,0.867,1}}23.78                    & {\cellcolor[rgb]{0.757,0.867,1}}22.56                    & {\cellcolor[rgb]{0.757,0.867,1}}17.68                    & {\cellcolor[rgb]{0.757,0.867,1}}14.63                    & {\cellcolor[rgb]{0.757,0.867,1}}17.07                     \\
                                              & AttentionCode\textsubscript{auto}                           & 23.78\textcolor{red}{\textsuperscript{+0.00}}            & 21.34\textcolor[rgb]{0,0.392,0}{\textsuperscript{-1.22}} & 17.68\textcolor{red}{\textsuperscript{+0.00}}            & 14.63\textcolor{red}{\textsuperscript{+0.00}}            & 16.46\textcolor[rgb]{0,0.392,0}{\textsuperscript{-0.61}}  \\
                                              & AttentionCoder\textsubscript{human}                          & 23.78\textcolor{red}{\textsuperscript{+0.00}}            & 22.56\textcolor{red}{\textsuperscript{+0.00}}            & 17.68\textcolor{red}{\textsuperscript{+0.0}}0            & 15.24\textcolor{red}{\textsuperscript{+0.61}}            & 16.46\textcolor[rgb]{0,0.392,0}{\textsuperscript{-0.61}}  \\ 
\hdashline
\multirow{3}{*}{WizardCoder 3B}               & {\cellcolor[rgb]{0.757,0.867,1}}No Attention & {\cellcolor[rgb]{0.757,0.867,1}}32.32                    & {\cellcolor[rgb]{0.757,0.867,1}}23.17                    & {\cellcolor[rgb]{0.757,0.867,1}}23.78                    & {\cellcolor[rgb]{0.757,0.867,1}}21.95                    & {\cellcolor[rgb]{0.757,0.867,1}}25.61                     \\
                                              & AttentionCode\textsubscript{auto}                           & 34.76\textcolor{red}{\textsuperscript{+2.44}}            & 26.83\textcolor{red}{\textsuperscript{+3.66}}            & 25.00\textcolor{red}{\textsuperscript{+1.22}}            & 20.73\textsuperscript{\textcolor[rgb]{0,0.392,0}{-1.22}} & 25.00\textcolor[rgb]{0,0.392,0}{\textsuperscript{-0.61}}  \\
                                              & AttentionCoder\textsubscript{human}                          & 34.14\textcolor{red}{\textsuperscript{+1.82}}            & 26.22\textcolor{red}{\textsuperscript{+3.05}}            & 25.00\textcolor{red}{\textsuperscript{+1.22}}            & 20.73\textcolor[rgb]{0,0.392,0}{\textsuperscript{-1.22}} & 25.00\textcolor[rgb]{0,0.392,0}{\textsuperscript{-0.61}}  \\ 
\hdashline
\multirow{3}{*}{WizardCoder 7B}               & {\cellcolor[rgb]{0.757,0.867,1}}No Attention & {\cellcolor[rgb]{0.757,0.867,1}}51.22                    & {\cellcolor[rgb]{0.757,0.867,1}}41.46                    & {\cellcolor[rgb]{0.757,0.867,1}}39.02                    & {\cellcolor[rgb]{0.757,0.867,1}}38.41                    & {\cellcolor[rgb]{0.757,0.867,1}}35.37                     \\
                                              & AttentionCode\textsubscript{auto}                           & 52.44\textcolor{red}{\textsuperscript{+1.22}}            & 46.34\textcolor{red}{\textsuperscript{+4.88}}            & 40.85\textcolor{red}{\textsuperscript{+1.83}}            & 40.85\textcolor{red}{\textsuperscript{+2.44}}            & 41.46\textcolor{red}{\textsuperscript{+6.09}}             \\
                                              & AttentionCoder\textsubscript{human}                          & 52.44\textcolor{red}{\textsuperscript{+1.22}}            & 43.90\textcolor{red}{\textsuperscript{+2.44}}            & 40.85\textcolor{red}{\textsuperscript{+1.83}}            & 40.24\textcolor{red}{\textsuperscript{+1.83}}            & 42.07\textcolor{red}{\textsuperscript{+6.70}}             \\ 
\hdashline
\multirow{3}{*}{WizardCoder 15B}              & {\cellcolor[rgb]{0.757,0.867,1}}No Attention & {\cellcolor[rgb]{0.757,0.867,1}}56.10                    & {\cellcolor[rgb]{0.757,0.867,1}}46.34                    & {\cellcolor[rgb]{0.757,0.867,1}}43.29                    & {\cellcolor[rgb]{0.757,0.867,1}}40.85                    & {\cellcolor[rgb]{0.757,0.867,1}}42.07                     \\
                                              & AttentionCode\textsubscript{auto}                           & 59.15\textcolor{red}{\textsuperscript{+3.05}}            & 52.44\textcolor{red}{\textsuperscript{+6.10}}            & 45.73\textcolor{red}{\textsuperscript{+2.44}}            & 42.68\textcolor{red}{\textsuperscript{+1.83}}            & 43.29\textcolor{red}{\textsuperscript{+1.22}}             \\
                                              & AttentionCoder\textsubscript{human}                          & 59.15\textcolor{red}{\textsuperscript{+3.05}}            & 51.83\textcolor{red}{\textsuperscript{+5.49}}            & 45.73\textcolor{red}{\textsuperscript{+2.44}}            & 43.29\textcolor{red}{\textsuperscript{+2.44}}            & 42.68\textcolor{red}{\textsuperscript{+0.61}}             \\ 
\hdashline
\multirow{3}{*}{CodeLlama 7B}                 & {\cellcolor[rgb]{0.757,0.867,1}}No Attention & {\cellcolor[rgb]{0.757,0.867,1}}32.93                         & {\cellcolor[rgb]{0.757,0.867,1}}28.05                         & {\cellcolor[rgb]{0.757,0.867,1}}25.00                         & {\cellcolor[rgb]{0.757,0.867,1}}23.78                         & {\cellcolor[rgb]{0.757,0.867,1}}21.95                          \\
                                              & AttentionCode\textsubscript{auto}                           & 37.20\textcolor{red}{\textsuperscript{+4.27}}            & 32.93\textcolor{red}{\textsuperscript{+4.87}}            & 32.93\textcolor{red}{\textsuperscript{+7.93}}            & 32.93\textcolor{red}{\textsuperscript{+9.15}}            & 28.66\textsuperscript{\textcolor{red}{+6.71}}                                                           \\
                                              & AttentionCoder\textsubscript{human}                           & 37.80\textcolor{red}{\textsuperscript{+4.87}}            & 33.54\textcolor{red}{\textsuperscript{+5.49}}            & 32.93\textcolor{red}{\textsuperscript{+7.93}}            & 33.54\textcolor{red}{\textsuperscript{+9.76}}            & 26.22\textcolor{red}{\textsuperscript{+4.27}}                                                           \\ 
\hdashline
\multirow{3}{*}{CodeLlama 13B}                & {\cellcolor[rgb]{0.757,0.867,1}}No Attention & {\cellcolor[rgb]{0.757,0.867,1}}40.85                         & {\cellcolor[rgb]{0.757,0.867,1}}39.02                         & {\cellcolor[rgb]{0.757,0.867,1}}35.37                         & {\cellcolor[rgb]{0.757,0.867,1}}34.15                         & {\cellcolor[rgb]{0.757,0.867,1}}32.32                          \\
                                              & AttentionCode\textsubscript{auto}                           & 44.51\textcolor{red}{\textsuperscript{+3.66}}            & 42.07\textcolor{red}{\textsuperscript{+3.05}}            & 37.80\textcolor{red}{\textsuperscript{+2.43}}            & 38.41\textcolor{red}{\textsuperscript{+4.26}}            & 34.15\textcolor{red}{\textsuperscript{+1.83}}                                     \\
                                              & AttentionCoder\textsubscript{human}                           & 45.12\textcolor{red}{\textsuperscript{+4.27}}            & 42.07\textcolor{red}{\textsuperscript{+3.05}}            & 37.80\textcolor{red}{\textsuperscript{+2.43}}            & 37.80\textcolor{red}{\textsuperscript{+3.66}}            & 35.37\textcolor{red}{\textsuperscript{+3.05}}                                                      \\
\hdashline
\multicolumn{1}{c|}{\multirow{5}{*}{GPT-3.5}} & {\cellcolor[rgb]{0.757,0.867,1}}No Attention & {\cellcolor[rgb]{0.757,0.867,1}}71.34                    & {\cellcolor[rgb]{0.757,0.867,1}}66.46                    & {\cellcolor[rgb]{0.757,0.867,1}}62.20                    & {\cellcolor[rgb]{0.757,0.867,1}}64.02                    & {\cellcolor[rgb]{0.757,0.867,1}}62.20                     \\
\multicolumn{1}{c|}{}                         & CoT Prompting                                & 68.29\textsuperscript{\textcolor[rgb]{0,0.392,0}{-3.05}} & 60.98\textsuperscript{\textcolor[rgb]{0,0.392,0}{-5.48}} & 56.71\textcolor[rgb]{0,0.392,0}{\textsuperscript{-5.49}} & 64.63\textcolor{red}{\textsuperscript{+0.61}}            & 55.49\textcolor[rgb]{0,0.392,0}{\textsuperscript{-6.71}}  \\
\multicolumn{1}{c|}{}                         & Reflexion                                    & 65.17\textsuperscript{\textcolor[rgb]{0,0.392,0}{-6.17}} & 60.37\textsuperscript{\textcolor[rgb]{0,0.392,0}{-6.09}} & 55.44\textcolor[rgb]{0,0.392,0}{\textsuperscript{-6.76}} & 56.10\textcolor[rgb]{0,0.392,0}{\textsuperscript{-7.92}} & 59.15\textcolor[rgb]{0,0.392,0}{\textsuperscript{-2.05}}  \\
\cdashline{2-7}
\multicolumn{1}{c|}{}                         & AttentionCode\textsubscript{auto}                           & 75.00\textcolor{red}{\textsuperscript{+3.66}}            & 74.39\textcolor{red}{\textsuperscript{+7.93}}            & 66.46\textcolor{red}{\textsuperscript{+4.26}}            & 67.68\textsuperscript{\textcolor{red}{+3.66}}            & 62.80\textcolor{red}{\textsuperscript{+0.60}}             \\
\multicolumn{1}{c|}{}                         & AttentionCoder\textsubscript{human}                          & 79.27\textcolor{red}{\textsuperscript{+7.93}}            & 76.83\textcolor{red}{\textsuperscript{+10.37}}           & 70.12\textcolor{red}{\textsuperscript{+7.92}}            & 70.73\textcolor{red}{\textsuperscript{+6.71}}            & 69.21\textcolor{red}{\textsuperscript{+7.01}}             \\ 
\hdashline
\multicolumn{1}{c|}{\multirow{5}{*}{GPT-4}}   & {\cellcolor[rgb]{0.757,0.867,1}}No Attention & {\cellcolor[rgb]{0.757,0.867,1}}81.20                    & {\cellcolor[rgb]{0.757,0.867,1}}76.82                    & {\cellcolor[rgb]{0.757,0.867,1}}70.73                    & {\cellcolor[rgb]{0.757,0.867,1}}70.19                    & {\cellcolor[rgb]{0.757,0.867,1}}73.17                     \\
\multicolumn{1}{c|}{}                         & CoT Prompting                                & 84.15\textcolor{red}{\textsuperscript{+2.95}}            & 81.10\textcolor{red}{\textsuperscript{+4.28}}            & 75.61\textcolor{red}{\textsuperscript{+4.88}}            & 75.00\textcolor{red}{\textsuperscript{+4.81}}            & 73.17\textcolor{red}{\textsuperscript{+0.00}}             \\
\multicolumn{1}{c|}{}                         & Reflexion                                    & 79.53\textsuperscript{\textcolor[rgb]{0,0.392,0}{-1.67}} & 80.49\textcolor{red}{\textsuperscript{+3.67}}            & 75.00\textcolor{red}{\textsuperscript{+4.27}}            & 72.56\textcolor{red}{\textsuperscript{+2.37}}            & 73.78\textcolor{red}{\textsuperscript{+0.61}}                               \\
\cdashline{2-7}
\multicolumn{1}{c|}{}                         & AttentionCode\textsubscript{auto}                           & 82.32\textcolor{red}{\textsuperscript{+1.12}}            & 78.05\textcolor{red}{\textsuperscript{+1.23}}            & 76.83\textcolor{red}{\textsuperscript{+6.10}}            & 75.00\textcolor{red}{\textsuperscript{+4.81}}            & 76.83\textcolor{red}{\textsuperscript{+3.66}}             \\
\multicolumn{1}{c|}{}                         & AttentionCoder\textsubscript{human}                          & 82.32\textcolor{red}{\textsuperscript{+1.12}}            & 78.05\textcolor{red}{\textsuperscript{+1.23}}            & 77.44\textcolor{red}{\textsuperscript{+6.71}}            & 75.00\textcolor{red}{\textsuperscript{+4.81}}            & 76.83\textcolor{red}{\textsuperscript{+3.66}}             \\ 
\bottomrule
\end{tabular}
\caption{Pass$@1$(\%) results on the \multinlh: \textcolor{blue}{blue} background is No Attention setting (baseline).}
\label{table: main results}
\end{table*}

\subsection{Effectiveness Validation}

To find out whether our framework can boost code LLMs on code generation in multiple-natural-language-oriented code generation scenario, we compare the performance of the base models on \multinlh in the three following settings:
(1) \textit{No Attention}: prompting models without the incorporation of extracted key phrases (\attention),
(2) \textit{AttentionCoder\textsubscript{auto}}: prompting models with key phrases extracted automatically through traditional NLP analysis tools,
(3) \textit{AttentionCoder\textsubscript{human}}: prompting models with key phrases extracted manually by human experts.

\begin{figure*}[ht]
    \centering
    \includegraphics[width=1\linewidth]{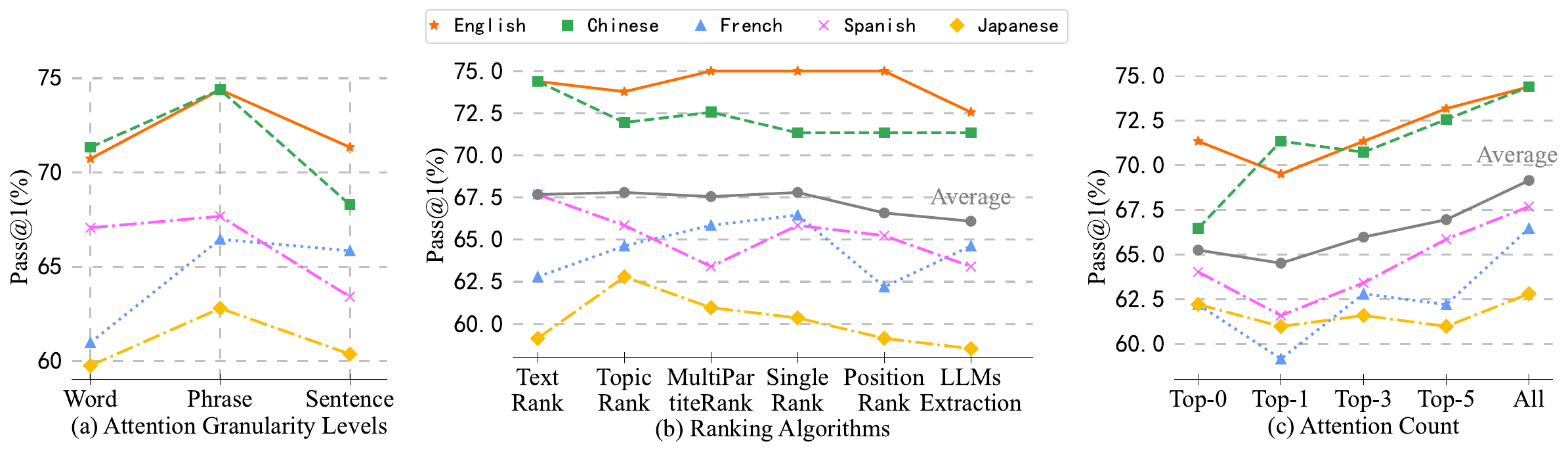}
    \caption{The result of influence factors on GPT-3.5: Attention Granularity Levels, Ranking Algorithms, Attention Count; ``Top-\textbf{x}'' in (c) means providing LLMs with \textbf{x} key phrases extracted by \attentionextractor, ``Top-0'' donates baseline (No Attention), ``All'' means providing LLMs all extracted \textit{Attention}.}
    \label{fig: factors}
\end{figure*}

Table \ref{table: main results} shows the performance of baseline models on \multinlh under the aforementioned three settings.
By comparing the results under \textit{No Attention} setting and the other two settings, it can be deduced that prompting \attention on existing code LLMs is capable of leading to better performance.
Particularly, with Chinese code comment, the \textit{AttentionCoder\textsubscript{auto}} setting of GPT-3.5 achieves a large absolute improvement of $7.93\%$ on pass$@1$, compared with the \textit{No Attention} setting.
This observation suggests that highlighting the extracted \textit{Attention} in prompt can unlock the potential of code LLMs to better understand programming problem.
Furthermore, by comparing the results under the \textit{AttentionCoder\textsubscript{auto}} and \textit{AttentionCoder\textsubscript{human}} setting, it can be observed that humans in the loop can usually further improve the code generation performance of code LLMs. 
For example, for GPT-3.5, the human setting can lead to an absolute improvement of $10.37\%$ on pass$@1$ compared to its baseline under the Chinese scenario.
A plausible explanation for this observation is that humans, unlike traditional NLP tools, excel in identifying and extracting pivotal phrases from natural language, thereby enabling code LLMs to better grasp the essence of the given natural language input.
We also compare two prompt engineering approaches including CoT~\cite{cot,cot_code} and Reflexion~\cite{shinn2023reflexion} with our framework on GPT-3.5 and GPT-4. 
And we observe that our approach consistently surpasses them across all languages. 
We hypothesize that this is primarily due to the erroneous cascading within CoT and Reflexion, further leading to a worse performance than our framework.
Interestingly, we observe that our baselines, CoT and Reflexion, show a declining trend on GPT-3.5 while exhibiting an upward trend on the more powerful GPT-4 model. This suggests that they demand more stringent foundational capabilities from the models.
Overall, all the above results prove the effectiveness of our approach to promote model understanding of code requirements.

\subsection{Ablation Study}

To delve into the effectiveness of our framework, we are keen on what factors can influence its performance.
We initially explore the impact of different \attention granularity levels on performance, including words, phrases, and sentences.
Following that, we investigate the impact of various ranking algorithms on performance, encompassing TextRank~\cite{mihalcea2004textrank}, TopicRank~\cite{bougouin2013topicrank}, MultiPartiteRank~\cite{boudin2018unsupervised}, PositionRank~\cite{florescu2017positionrank}, SingleRank~\cite{wan2008single}, and 
LLMs extraction.
Subsequently, we analyze the impact of extracted phrase count on performance.
Finally, we assess the effectiveness of our framework on models with varying parameter sizes. 
Owing to space limitations, more detailed analysis are provided in Appendix~\ref{apx: attention type}.

\paragraph{\attention Granularity Levels}
\label{para: attention form}
Technically, attention can be applied at different levels of granularity, including words, phrases, or sentences. 
Therefore, we compare these levels and the results are presented in Figure~\ref{fig: factors} (a).
It is observed that the performance of phrase-based attention surpasses that of word-based and sentence-based attention. 
A reasonable conjecture is that phrases more effectively highlight the key information, in contrast to words and sentences which may miss these focal points due to being overly brief or excessively lengthy.
Consequently, for our default experiment, we adopt phrase-based \attention.

\paragraph{Ranking Algorithms}
\label{rank algorithms}
Extracting key phrases from natural language and ranking them in importance is a very important step in our framework. Therefore, we explore the effect of different ranking algorithms on \multinlh. Specifically, these include TextRank, TopicRank, MultiPartiteRank, SingleRank, and PositionRank. The results of these algorithms are shown in Figure~\ref{fig: factors} (b). We find that these rank algorithms perform almost the same. This shows that our framework is insensitive to ranking algorithms.
In addition to using traditional NLP analysis tools, we also use OpenAI's GPT-3.5 to automatically extract and rank the attention. Details about the prompt can be seen in Appendix~\ref{apx:llms extraction}. The results show that LLMs extraction is less effective, further demonstrating the necessity and good prospect of combining with traditional NLP tools for LLMs.

\paragraph{\attention Count}
We would like to explore how many phrases should be prompted to LLMs for appropriate \attention, and the results are displayed in Figure~\ref{fig: factors} (c).
The results show that for almost all languages, performance initially declines and then increases as the count of phrases increases, peaking when all phrases are prompted.
This suggests that prompting with too few phrases might prevent the LLM from grasping the key information.
Under this context, we are intrigued to determine whether using the entire programming problem as \attention, without extracting phrases, could improve performance.
Results in Figure~\ref{fig: factors} (a) show that this setting (``Sentence'') leads to a decrease in performance, further underscoring the necessity of using phrases as \attention.

\begin{figure}[h]
    \centering
    \includegraphics[width=0.75\linewidth]{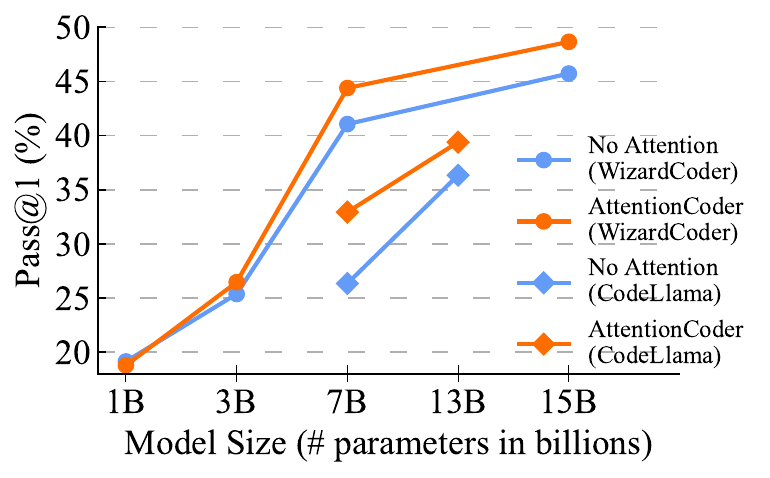}
    \caption{The results of model size: ordinate value is the average of five natural language results on Pass$@1$.}
    \label{fig: model size}
\end{figure}

\paragraph{Model Size}
We display the performance of WizardCoder and CodeLlama of various sizes in Figure~\ref{fig: model size}.
We observe that our framework has a threshold for model size.
For instance, WizardCoder $1$B achieves no improvement in pass$@1$ across all languages, while WizardCoder $7$B brought a $3.4$\% average improvement.
This suggests that by prompting \attention, the powerful models can obtain obvious improvement, while small models may not.

\begin{figure*}[ht]
    \centering
    \includegraphics[width=0.8\linewidth]{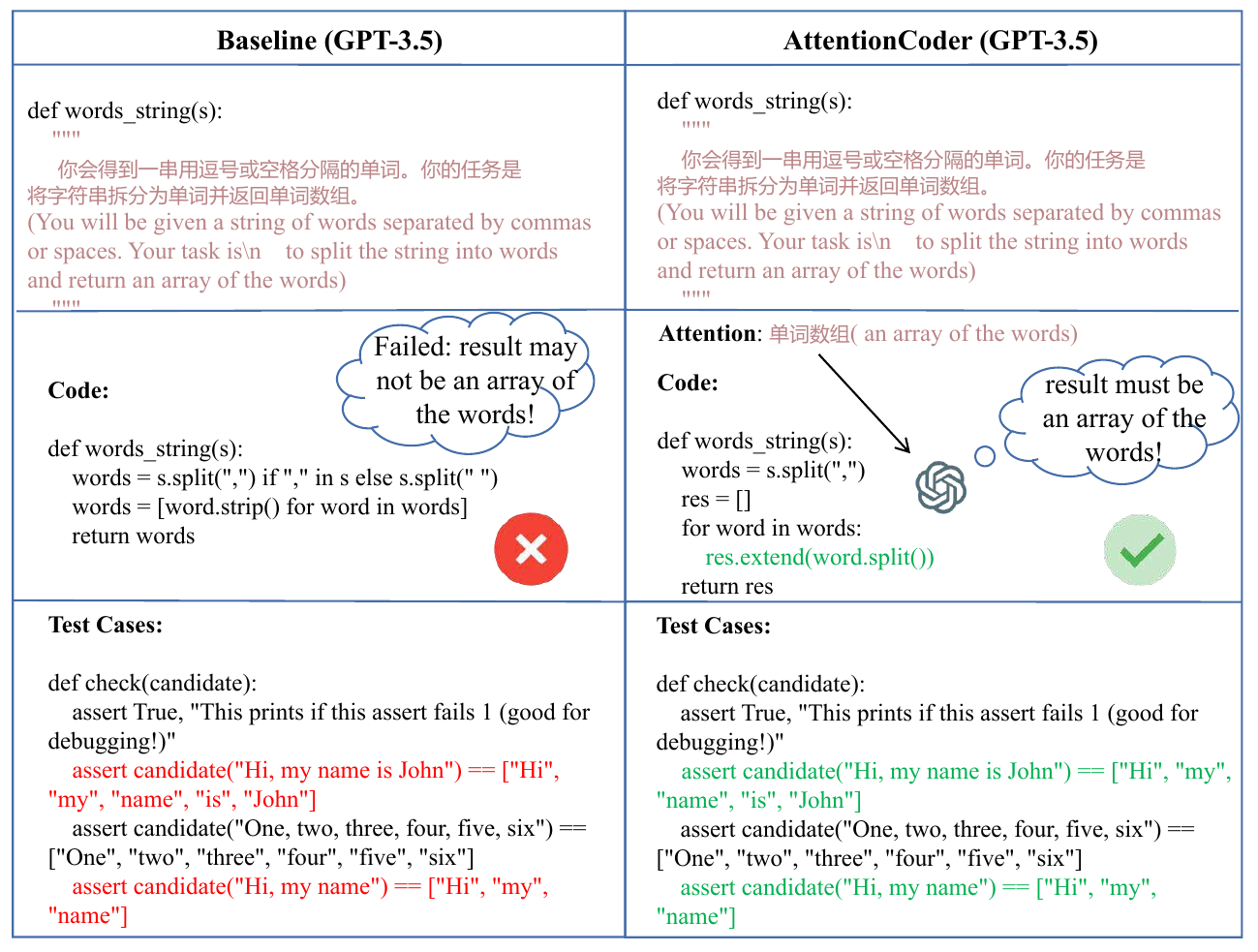}
    \caption{One concrete case from the \multinlh benchmark.}
    \label{fig:caseStudy}
\end{figure*}

\subsection{Transferability Evaluation}
\label{sec:transferability_evaluation}
Intuitively, as a means to assist the model in understanding natural language tasks, our framework has the potential to excel in various tasks involving natural-language context.
Consequently, we are curious about whether \textit{Attention} helps in different tasks across various domains.
To declare our framework's transferability, we extend it to three other tasks:
(1) code generation of multiple programming languages, evaluated using the MultiPL-E~\cite{multipl-e} benchmark.
(2) code translation, evaluated using the HumanEval-X~\cite{codegeex} benchmark, from which we specifically select two sub-tasks related to code translation: py2java and py2c++.
(3) mathematical reasoning, evaluated using the GSM8k~\cite{gsm8k} benchmark.
For these tasks, we implement our framework by reusing the \attentionextractor and landing \attentioncoder with versatile OpenAI's GPT-3.5.
As for the three different downstream tasks, we utilize different templates to import \textit{Attention} and prompt the model. More details about these templates can be seen in Appendix~\ref{apx:other_task_template}. 

The evaluation results on the three tasks are presented in Table~\ref{table: other tasks}.
For code generation of multiple programming languages, our framework implemented based on GPT-3.5 obtains different degrees of improvement on different programming languages.
For example, it leads to an absolute improvement of $6.2\%$ on pass$@1$ of the JavaScript code generation tasks compared to the performance of GPT-3.5 without our framework.
This indicates the efficiency of our framework in code generation of multiple programming languages.
For code translation tasks, with our framework, GPT-3.5 achieves a comprehensive enhancement on both of the tasks, which further shows the effectiveness of our framework.
For instance, our framework produces an absolute pass$@1$ improvement of $3.05\%$ for py2java and that of $11.58\%$ for py2c++.
It implies that the improvement brought about by our framework in code translation depends greatly on the involved programming languages.
Actually, this observation is also consistent with the results on code generation of multiple programming languages, where the improvement also varies according to the programming language used to generate codes.
For mathematical reasoning, our framework only derives a tiny improvement on GSM8K.
One possible reason is that GSM8K's task descriptions, characterized by structured and simple sentences, differ significantly in complexity from those in HumanEval, \multinlh, HumanEval-X, and MultiPL-E, which tend to be more intricate.
Thus, our framework may excel in handling relatively complex natural language scenarios, with less pronounced advantages in simpler tasks.
In summary, our framework consistently demonstrates advantages across various tasks, showing the potential of transferability.

Of note, the experiments to evaluate the transferability of our framework only focus on the English version.
However, we believe the improvement brought about by our framework could be extended to other natural languages, especially considering the conclusions already drawn from RQ1.

\begin{table}[t]
\small
\centering
\begin{tblr}{
  cell{1}{1} = {c=2}{c},
  cell{2}{1} = {r=3}{},
  cell{5}{1} = {r=2}{},
  cell{2}{4} = {c},
  cell{3}{4} = {c},
  cell{4}{4} = {c},
  cell{5}{4} = {c},
  cell{6}{4} = {c},
  cell{7}{4} = {c},
  cell{2}{3} = {c},
  cell{3}{3} = {c},
  cell{4}{3} = {c},
  cell{5}{3} = {c},
  cell{6}{3} = {c},
  cell{7}{3} = {c},
  vline{3} = {1-7}{},
  vline{2} = {2-7}{},
  hline{1-2,5,7,8} = {-}{},
}
\textbf{Evaluation Setting} &                  & \textbf{Baseline} & {\textbf{AttentionCoder}}      \\
MultiPL-E     & go               & 27.92             & \textbf{\textbf{\textbf{\textbf{31.17}}}} \\
              & js               & 50.93             & \textbf{\textbf{\textbf{\textbf{57.14}}}} \\
              & java             & 48.10             & \textbf{\textbf{\textbf{\textbf{49.37}}}} \\
HumanEval-X~  & py2java          & 76.83             & \textbf{\textbf{79.88}}                   \\
              & py2c++           & 51.83             & \textbf{\textbf{63.41}}                   \\
GSM8K         & -                & 74.60             & \textbf{\textbf{74.90}}                   
\end{tblr}
\caption{Pass$@1$(\%) results on GPT-3.5 of different tasks.}
\label{table: other tasks}
\end{table}

\subsection{Case Study}
After extensive experiments, we can conclude that introducing \attention can help LLMs to better understand the programming context.
One practical case is displayed in Figure~\ref{fig:caseStudy}.
For the baseline without \attention, the generated code cannot pass all test cases.
The primary reason is that it ignores the ``\texttt{an array of the words}'' specified in the requirements.
But when we add this phase into the prompt to highlight, code LLMs could pay more attention to the return type, thereby, generating the correct code.

\section{Related Work}
\subsection{Code Large Language Model}

Code large language models~\cite{codet,cert,liu2023uncovering,apicoder,repocoder} have attracted widespread attention in both industry and academia.
As a milestone, Codex~\cite{codex} demonstrates great code generation abilities. 
After that, a plenty of code LLMs released, such as PolyCoder~\cite{polycoder}, CodeParrot~\cite{codeparrot}, AlphaCode~\cite{alphacode}, PaLM-Coder~\cite{pangu-coder}, StarCoder~\cite{starcoder}, CodeLlama~\cite{codellama}, WizardCoder~\cite{wizardcoder}. 
Some efforts~\cite{codegeex,multipl-e} have evaluated code LLMs from the perspective of multiple programming languages. 
However, few works focus on multiple-natural-language-oriented code generation. 
In this paper, we would like to explore the natural language capabilities of code LLMs by proposing a prompt engineering framework to unleash the potential of code LLMs.

\subsection{Phrase Extraction}
Phrase extraction is a traditional NLP task, which involves extracting key phrases~\cite{turney2000learning} from given natural language text. Many approaches have been proposed, which can be divided into two types: supervised manners, such as deep key phrase generation~\cite{meng2017deep} and LTR~\cite{burges2005learning}, and unsupervised ones, such as TextRank~\cite{mihalcea2004textrank}, TopicRank~\cite{bougouin2013topicrank}, MultiPartiteRank~\cite{boudin2018unsupervised}, SingleRank~\cite{wan2008single}, and PositionRank~\cite{florescu2017positionrank}. 
In this paper, we regard these unsupervised methods are easy yet effective to extract key phrases from given natural language code descriptions. 
The extracted key phrases can be prompted to code LLMs to enhance natural language comprehension.

\section{Conclusion}
This paper proposes a simple yet effective framework, including \attentionextractor and \attentioncoder, for improving the natural language capabilities of code LLMs.
The essence of this framework lies in its seamless integration of LLMs with traditional NLP tools.
To thoroughly evaluate our framework, we craft a multiple-natural-language-oriented benchmark, namely \multinlh.
Finally, extensive experiments demonstrate the effectiveness and transferability of our framework.
In future work, we would like to investigate how to implement our framework during the pre-training and fine-tuning phase, not just at inference.

\section*{Limitations}
Our framework also encompasses some limitations:
(1) Technically, our framework can be applied to any task requiring natural language understanding. 
However, validating our framework across all downstream tasks is impractical. 
Therefore, in Section~\ref{sec:transferability_evaluation}, we select three popular downstream tasks for detailed analysis. 
Similarly, our framework is designed to process various natural languages. However, it is impractical to craft a dataset for every natural language in the world, considering a total of upwards to $7,000$\footnote{\url{https://en.wikipedia.org/wiki/Lists_of_languages}}.
Hence, we choose five representative languages (English, Chinese, French, Spanish, and Japanese) for which we create benchmarks in code generation.
(2) Our framework is based on prompt engineering, and it costs a large amount of computing resources. 
Therefore, we will open source all our code and data to alleviate this issue.

\section*{Acknowledgments}
This research was supported by the National Key Research and Development Program of China, under Grant No. 2022ZD0120201 - ``Unified Representation and Knowledge Graph Construction for Science Popularization Resources''.

\bibliography{anthology,custom}

\begin{thebibliography}{38}
\expandafter\ifx\csname natexlab\endcsname\relax\def\natexlab#1{#1}\fi

\bibitem[{Boudin(2018)}]{boudin2018unsupervised}
Florian Boudin. 2018.
\newblock Unsupervised keyphrase extraction with multipartite graphs.
\newblock \emph{arXiv preprint arXiv:1803.08721}.

\bibitem[{Bougouin et~al.(2013)Bougouin, Boudin, and Daille}]{bougouin2013topicrank}
Adrien Bougouin, Florian Boudin, and B{\'e}atrice Daille. 2013.
\newblock Topicrank: Graph-based topic ranking for keyphrase extraction.
\newblock In \emph{International joint conference on natural language processing (IJCNLP)}, pages 543--551.

\bibitem[{Burges et~al.(2005)Burges, Shaked, Renshaw, Lazier, Deeds, Hamilton, and Hullender}]{burges2005learning}
Chris Burges, Tal Shaked, Erin Renshaw, Ari Lazier, Matt Deeds, Nicole Hamilton, and Greg Hullender. 2005.
\newblock Learning to rank using gradient descent.
\newblock In \emph{Proceedings of the 22nd international conference on Machine learning}, pages 89--96.

\bibitem[{Cassano et~al.(2022)Cassano, Gouwar, Nguyen, Nguyen, Phipps-Costin, Pinckney, Yee, Zi, Anderson, Feldman, Guha, Greenberg, and Jangda}]{multipl-e}
Federico Cassano, John Gouwar, Daniel Nguyen, Sy~Duy Nguyen, Luna Phipps-Costin, Donald Pinckney, Ming-Ho Yee, Yangtian Zi, Carolyn~Jane Anderson, Molly~Q. Feldman, Arjun Guha, Michael Greenberg, and Abhinav Jangda. 2022.
\newblock A scalable and extensible approach to benchmarking nl2code for 18 programming languages.
\newblock \emph{ArXiv}, abs/2208.08227.

\bibitem[{Chen et~al.(2023)Chen, Zhang, Nguyen, Zan, Lin, Lou, and Chen}]{codet}
Bei Chen, Fengji Zhang, Anh Nguyen, Daoguang Zan, Zeqi Lin, Jian-Guang Lou, and Weizhu Chen. 2023.
\newblock {CodeT}: Code generation with generated tests.
\newblock In \emph{The Eleventh International Conference on Learning Representations}.

\bibitem[{Chen et~al.(2021)Chen, Tworek, Jun, Yuan, Ponde, Kaplan, Edwards, Burda, Joseph, Brockman, Ray, Puri, Krueger, Petrov, Khlaaf, Sastry, Mishkin, Chan, Gray, Ryder, Pavlov, Power, Kaiser, Bavarian, Winter, Tillet, Such, Cummings, Plappert, Chantzis, Barnes, Herbert-Voss, Guss, Nichol, Babuschkin, Balaji, Jain, Carr, Leike, Achiam, Misra, Morikawa, Radford, Knight, Brundage, Murati, Mayer, Welinder, McGrew, Amodei, McCandlish, Sutskever, and Zaremba}]{codex}
Mark Chen, Jerry Tworek, Heewoo Jun, Qiming Yuan, Henrique Ponde, Jared Kaplan, Harrison Edwards, Yura Burda, Nicholas Joseph, Greg Brockman, Alex Ray, Raul Puri, Gretchen Krueger, Michael Petrov, Heidy Khlaaf, Girish Sastry, Pamela Mishkin, Brooke Chan, Scott Gray, Nick Ryder, Mikhail Pavlov, Alethea Power, Lukasz Kaiser, Mohammad Bavarian, Clemens Winter, Philippe Tillet, Felipe~Petroski Such, David~W. Cummings, Matthias Plappert, Fotios Chantzis, Elizabeth Barnes, Ariel Herbert-Voss, William~H. Guss, Alex Nichol, Igor Babuschkin, S.~Arun Balaji, Shantanu Jain, Andrew Carr, Jan Leike, Joshua Achiam, Vedant Misra, Evan Morikawa, Alec Radford, Matthew~M. Knight, Miles Brundage, Mira Murati, Katie Mayer, Peter Welinder, Bob McGrew, Dario Amodei, Sam McCandlish, Ilya Sutskever, and Wojciech Zaremba. 2021.
\newblock Evaluating large language models trained on code.
\newblock \emph{ArXiv}, abs/2107.03374.

\bibitem[{Christopoulou et~al.(2022)Christopoulou, Lampouras, Gritta, Zhang, Guo, Li, Zhang, Xiao, Shen, Li, Yu, yu~Yan, Zhou, Wang, Ma, Iacobacci, Wang, Liang, Wei, Jiang, Wang, and Liu}]{pangu-coder}
Fenia Christopoulou, Gerasimos Lampouras, Milan Gritta, Guchun Zhang, Yinpeng Guo, Zhong-Yi Li, Qi~Zhang, Meng Xiao, Bo~Shen, Lin Li, Hao Yu, Li~yu~Yan, Pingyi Zhou, Xin Wang, Yu~Ma, Ignacio Iacobacci, Yasheng Wang, Guangtai Liang, Jia Wei, Xin Jiang, Qianxiang Wang, and Qun Liu. 2022.
\newblock {PanGu-Coder}: Program synthesis with function-level language modeling.
\newblock \emph{ArXiv}, abs/2207.11280.

\bibitem[{Cobbe et~al.(2021)Cobbe, Kosaraju, Bavarian, Chen, Jun, Kaiser, Plappert, Tworek, Hilton, Nakano et~al.}]{gsm8k}
Karl Cobbe, Vineet Kosaraju, Mohammad Bavarian, Mark Chen, Heewoo Jun, Lukasz Kaiser, Matthias Plappert, Jerry Tworek, Jacob Hilton, Reiichiro Nakano, et~al. 2021.
\newblock Training verifiers to solve math word problems.
\newblock \emph{arXiv preprint arXiv:2110.14168}.

\bibitem[{Florescu and Caragea(2017)}]{florescu2017positionrank}
Corina Florescu and Cornelia Caragea. 2017.
\newblock Positionrank: An unsupervised approach to keyphrase extraction from scholarly documents.
\newblock In \emph{Proceedings of the 55th annual meeting of the association for computational linguistics (volume 1: long papers)}, pages 1105--1115.

\bibitem[{Huggingface(2021)}]{codeparrot}
Huggingface. 2021.
\newblock {Training CodeParrot from Scratch}.
\newblock \url{https://huggingface.co/blog/codeparrot}.

\bibitem[{Jia~Li and Jin(2023)}]{scot}
Yongmin~Li Jia~Li, Ge~Li and Zhi Jin. 2023.
\newblock Structured chain-of-thought prompting for code generation.

\bibitem[{Kalyuga(2011)}]{kalyuga2011cognitive}
Slava Kalyuga. 2011.
\newblock Cognitive load theory: How many types of load does it really need?
\newblock \emph{Educational Psychology Review}, 23:1--19.

\bibitem[{Li et~al.(2023{\natexlab{a}})Li, Li, Li, and Jin}]{cot_code}
Jia Li, Ge~Li, Yongmin Li, and Zhi Jin. 2023{\natexlab{a}}.
\newblock Structured chain-of-thought prompting for code generation.
\newblock \emph{arXiv preprint arXiv:2305.06599}.

\bibitem[{Li et~al.(2023{\natexlab{b}})Li, Allal, Zi, Muennighoff, Kocetkov, Mou, Marone, Akiki, Li, Chim, Liu, Zheltonozhskii, Zhuo, Wang, Dehaene, Davaadorj, Lamy-Poirier, Monteiro, Shliazhko, Gontier, Meade, Zebaze, Yee, Umapathi, Zhu, Lipkin, Oblokulov, Wang, Murthy, Stillerman, Patel, Abulkhanov, Zocca, Dey, Zhang, Fahmy, Bhattacharyya, Yu, Singh, Luccioni, Villegas, Kunakov, Zhdanov, Romero, Lee, Timor, Ding, Schlesinger, Schoelkopf, Ebert, Dao, Mishra, Gu, Robinson, Anderson, Dolan-Gavitt, Contractor, Reddy, Fried, Bahdanau, Jernite, Ferrandis, Hughes, Wolf, Guha, von Werra, and de~Vries}]{starcoder}
Raymond Li, Loubna~Ben Allal, Yangtian Zi, Niklas Muennighoff, Denis Kocetkov, Chenghao Mou, Marc Marone, Christopher Akiki, Jia Li, Jenny Chim, Qian Liu, Evgenii Zheltonozhskii, Terry~Yue Zhuo, Thomas Wang, Olivier Dehaene, Mishig Davaadorj, Joel Lamy-Poirier, João Monteiro, Oleh Shliazhko, Nicolas Gontier, Nicholas Meade, Armel Zebaze, Ming-Ho Yee, Logesh~Kumar Umapathi, Jian Zhu, Benjamin Lipkin, Muhtasham Oblokulov, Zhiruo Wang, Rudra Murthy, Jason Stillerman, Siva~Sankalp Patel, Dmitry Abulkhanov, Marco Zocca, Manan Dey, Zhihan Zhang, Nour Fahmy, Urvashi Bhattacharyya, Wenhao Yu, Swayam Singh, Sasha Luccioni, Paulo Villegas, Maxim Kunakov, Fedor Zhdanov, Manuel Romero, Tony Lee, Nadav Timor, Jennifer Ding, Claire Schlesinger, Hailey Schoelkopf, Jan Ebert, Tri Dao, Mayank Mishra, Alex Gu, Jennifer Robinson, Carolyn~Jane Anderson, Brendan Dolan-Gavitt, Danish Contractor, Siva Reddy, Daniel Fried, Dzmitry Bahdanau, Yacine Jernite, Carlos~Muñoz Ferrandis, Sean Hughes, Thomas Wolf, Arjun Guha, Leandro von
  Werra, and Harm de~Vries. 2023{\natexlab{b}}.
\newblock \href {http://arxiv.org/abs/2305.06161} {{StarCoder}: may the source be with you!}

\bibitem[{Li et~al.(2022)Li, Choi, Chung, Kushman, Schrittwieser, Leblond, Tom, Eccles, Keeling, Gimeno, Lago, Hubert, Choy, de, d’Autume, Babuschkin, Chen, Huang, Welbl, Gowal, Alexey, Cherepanov, Molloy, Mankowitz, Robson, Kohli, de, Freitas, Kavukcuoglu, and Vinyals}]{alphacode}
Yujia Li, David~H. Choi, Junyoung Chung, Nate Kushman, Julian Schrittwieser, R{\'e}mi Leblond, Tom, Eccles, James Keeling, Felix Gimeno, Agustin~Dal Lago, Thomas Hubert, Peter Choy, Cyprien de, Masson d’Autume, Igor Babuschkin, Xinyun Chen, Po-Sen Huang, Johannes Welbl, Sven Gowal, Alexey, Cherepanov, James Molloy, Daniel~Jaymin Mankowitz, Esme~Sutherland Robson, Pushmeet Kohli, Nando de, Freitas, Koray Kavukcuoglu, and Oriol Vinyals. 2022.
\newblock Competition-level code generation with alphacode.
\newblock \emph{Science}, 378:1092 -- 1097.

\bibitem[{Liu et~al.(2023)Liu, Chen, Gao, Su, Zhang, Zan, Lou, Chen, and Ho}]{liu2023uncovering}
Yan Liu, Xiaokang Chen, Yan Gao, Zhe Su, Fengji Zhang, Daoguang Zan, Jian-Guang Lou, Pin-Yu Chen, and Tsung-Yi Ho. 2023.
\newblock Uncovering and quantifying social biases in code generation.
\newblock \emph{arXiv preprint arXiv:2305.15377}.

\bibitem[{Luo et~al.(2023)Luo, Xu, Zhao, Sun, Geng, Hu, Tao, Ma, Lin, and Jiang}]{wizardcoder}
Ziyang Luo, Can Xu, Pu~Zhao, Qingfeng Sun, Xiubo Geng, Wenxiang Hu, Chongyang Tao, Jing Ma, Qingwei Lin, and Daxin Jiang. 2023.
\newblock {WizardCoder}: Empowering code large language models with evol-instruct.
\newblock \emph{arXiv preprint arXiv:2306.08568}.

\bibitem[{Meng et~al.(2017)Meng, Zhao, Han, He, Brusilovsky, and Chi}]{meng2017deep}
Rui Meng, Sanqiang Zhao, Shuguang Han, Daqing He, Peter Brusilovsky, and Yu~Chi. 2017.
\newblock Deep keyphrase generation.
\newblock \emph{arXiv preprint arXiv:1704.06879}.

\bibitem[{Mihalcea and Tarau(2004)}]{mihalcea2004textrank}
Rada Mihalcea and Paul Tarau. 2004.
\newblock Textrank: Bringing order into text.
\newblock In \emph{Proceedings of the 2004 conference on empirical methods in natural language processing}, pages 404--411.

\bibitem[{Name(2023)}]{phind}
Author's Name. 2023.
\newblock \href {https://www.phind.com/blog/code-llama-beats-gpt4} {Code llama beats gpt-4: A deep dive}.
\newblock Accessed: 2023-10-31.

\bibitem[{OpenAI et~al.(2023)OpenAI, :, Achiam, Adler, Agarwal, Ahmad, Akkaya, Aleman, Almeida, Altenschmidt, Altman, Anadkat, Avila, Babuschkin, Balaji, Balcom, Baltescu, Bao, Bavarian, Belgum, Bello, Berdine, Bernadett-Shapiro, Berner, Bogdonoff, Boiko, Boyd, Brakman, Brockman, Brooks, Brundage, Button, Cai, Campbell, Cann, Carey, Carlson, Carmichael, Chan, Chang, Chantzis, Chen, Chen, Chen, Chen, Chen, Chess, Cho, Chu, Chung, Cummings, Currier, Dai, Decareaux, Degry, Deutsch, Deville, Dhar, Dohan, Dowling, Dunning, Ecoffet, Eleti, Eloundou, Farhi, Fedus, Felix, Fishman, Forte, Fulford, Gao, Georges, Gibson, Goel, Gogineni, Goh, Gontijo-Lopes, Gordon, Grafstein, Gray, Greene, Gross, Gu, Guo, Hallacy, Han, Harris, He, Heaton, Heidecke, Hesse, Hickey, Hickey, Hoeschele, Houghton, Hsu, Hu, Hu, Huizinga, Jain, Jain, Jang, Jiang, Jiang, Jin, Jin, Jomoto, Jonn, Jun, Kaftan, Łukasz Kaiser, Kamali, Kanitscheider, Keskar, Khan, Kilpatrick, Kim, Kim, Kim, Kirchner, Kiros, Knight, Kokotajlo, Łukasz Kondraciuk,
  Kondrich, Konstantinidis, Kosic, Krueger, Kuo, Lampe, Lan, Lee, Leike, Leung, Levy, Li, Lim, Lin, Lin, Litwin, Lopez, Lowe, Lue, Makanju, Malfacini, Manning, Markov, Markovski, Martin, Mayer, Mayne, McGrew, McKinney, McLeavey, McMillan, McNeil, Medina, Mehta, Menick, Metz, Mishchenko, Mishkin, Monaco, Morikawa, Mossing, Mu, Murati, Murk, Mély, Nair, Nakano, Nayak, Neelakantan, Ngo, Noh, Ouyang, O'Keefe, Pachocki, Paino, Palermo, Pantuliano, Parascandolo, Parish, Parparita, Passos, Pavlov, Peng, Perelman, de~Avila Belbute~Peres, Petrov, de~Oliveira~Pinto, Michael, Pokorny, Pokrass, Pong, Powell, Power, Power, Proehl, Puri, Radford, Rae, Ramesh, Raymond, Real, Rimbach, Ross, Rotsted, Roussez, Ryder, Saltarelli, Sanders, Santurkar, Sastry, Schmidt, Schnurr, Schulman, Selsam, Sheppard, Sherbakov, Shieh, Shoker, Shyam, Sidor, Sigler, Simens, Sitkin, Slama, Sohl, Sokolowsky, Song, Staudacher, Such, Summers, Sutskever, Tang, Tezak, Thompson, Tillet, Tootoonchian, Tseng, Tuggle, Turley, Tworek, Uribe, Vallone,
  Vijayvergiya, Voss, Wainwright, Wang, Wang, Wang, Ward, Wei, Weinmann, Welihinda, Welinder, Weng, Weng, Wiethoff, Willner, Winter, Wolrich, Wong, Workman, Wu, Wu, Wu, Xiao, Xu, Yoo, Yu, Yuan, Zaremba, Zellers, Zhang, Zhang, Zhao, Zheng, Zhuang, Zhuk, and Zoph}]{gpt-4}
OpenAI, :, Josh Achiam, Steven Adler, Sandhini Agarwal, Lama Ahmad, Ilge Akkaya, Florencia~Leoni Aleman, Diogo Almeida, Janko Altenschmidt, Sam Altman, Shyamal Anadkat, Red Avila, Igor Babuschkin, Suchir Balaji, Valerie Balcom, Paul Baltescu, Haiming Bao, Mo~Bavarian, Jeff Belgum, Irwan Bello, Jake Berdine, Gabriel Bernadett-Shapiro, Christopher Berner, Lenny Bogdonoff, Oleg Boiko, Madelaine Boyd, Anna-Luisa Brakman, Greg Brockman, Tim Brooks, Miles Brundage, Kevin Button, Trevor Cai, Rosie Campbell, Andrew Cann, Brittany Carey, Chelsea Carlson, Rory Carmichael, Brooke Chan, Che Chang, Fotis Chantzis, Derek Chen, Sully Chen, Ruby Chen, Jason Chen, Mark Chen, Ben Chess, Chester Cho, Casey Chu, Hyung~Won Chung, Dave Cummings, Jeremiah Currier, Yunxing Dai, Cory Decareaux, Thomas Degry, Noah Deutsch, Damien Deville, Arka Dhar, David Dohan, Steve Dowling, Sheila Dunning, Adrien Ecoffet, Atty Eleti, Tyna Eloundou, David Farhi, Liam Fedus, Niko Felix, Simón~Posada Fishman, Juston Forte, Isabella Fulford, Leo Gao,
  Elie Georges, Christian Gibson, Vik Goel, Tarun Gogineni, Gabriel Goh, Rapha Gontijo-Lopes, Jonathan Gordon, Morgan Grafstein, Scott Gray, Ryan Greene, Joshua Gross, Shixiang~Shane Gu, Yufei Guo, Chris Hallacy, Jesse Han, Jeff Harris, Yuchen He, Mike Heaton, Johannes Heidecke, Chris Hesse, Alan Hickey, Wade Hickey, Peter Hoeschele, Brandon Houghton, Kenny Hsu, Shengli Hu, Xin Hu, Joost Huizinga, Shantanu Jain, Shawn Jain, Joanne Jang, Angela Jiang, Roger Jiang, Haozhun Jin, Denny Jin, Shino Jomoto, Billie Jonn, Heewoo Jun, Tomer Kaftan, Łukasz Kaiser, Ali Kamali, Ingmar Kanitscheider, Nitish~Shirish Keskar, Tabarak Khan, Logan Kilpatrick, Jong~Wook Kim, Christina Kim, Yongjik Kim, Hendrik Kirchner, Jamie Kiros, Matt Knight, Daniel Kokotajlo, Łukasz Kondraciuk, Andrew Kondrich, Aris Konstantinidis, Kyle Kosic, Gretchen Krueger, Vishal Kuo, Michael Lampe, Ikai Lan, Teddy Lee, Jan Leike, Jade Leung, Daniel Levy, Chak~Ming Li, Rachel Lim, Molly Lin, Stephanie Lin, Mateusz Litwin, Theresa Lopez, Ryan Lowe,
  Patricia Lue, Anna Makanju, Kim Malfacini, Sam Manning, Todor Markov, Yaniv Markovski, Bianca Martin, Katie Mayer, Andrew Mayne, Bob McGrew, Scott~Mayer McKinney, Christine McLeavey, Paul McMillan, Jake McNeil, David Medina, Aalok Mehta, Jacob Menick, Luke Metz, Andrey Mishchenko, Pamela Mishkin, Vinnie Monaco, Evan Morikawa, Daniel Mossing, Tong Mu, Mira Murati, Oleg Murk, David Mély, Ashvin Nair, Reiichiro Nakano, Rajeev Nayak, Arvind Neelakantan, Richard Ngo, Hyeonwoo Noh, Long Ouyang, Cullen O'Keefe, Jakub Pachocki, Alex Paino, Joe Palermo, Ashley Pantuliano, Giambattista Parascandolo, Joel Parish, Emy Parparita, Alex Passos, Mikhail Pavlov, Andrew Peng, Adam Perelman, Filipe de~Avila Belbute~Peres, Michael Petrov, Henrique~Ponde de~Oliveira~Pinto, Michael, Pokorny, Michelle Pokrass, Vitchyr Pong, Tolly Powell, Alethea Power, Boris Power, Elizabeth Proehl, Raul Puri, Alec Radford, Jack Rae, Aditya Ramesh, Cameron Raymond, Francis Real, Kendra Rimbach, Carl Ross, Bob Rotsted, Henri Roussez, Nick Ryder,
  Mario Saltarelli, Ted Sanders, Shibani Santurkar, Girish Sastry, Heather Schmidt, David Schnurr, John Schulman, Daniel Selsam, Kyla Sheppard, Toki Sherbakov, Jessica Shieh, Sarah Shoker, Pranav Shyam, Szymon Sidor, Eric Sigler, Maddie Simens, Jordan Sitkin, Katarina Slama, Ian Sohl, Benjamin Sokolowsky, Yang Song, Natalie Staudacher, Felipe~Petroski Such, Natalie Summers, Ilya Sutskever, Jie Tang, Nikolas Tezak, Madeleine Thompson, Phil Tillet, Amin Tootoonchian, Elizabeth Tseng, Preston Tuggle, Nick Turley, Jerry Tworek, Juan Felipe~Cerón Uribe, Andrea Vallone, Arun Vijayvergiya, Chelsea Voss, Carroll Wainwright, Justin~Jay Wang, Alvin Wang, Ben Wang, Jonathan Ward, Jason Wei, CJ~Weinmann, Akila Welihinda, Peter Welinder, Jiayi Weng, Lilian Weng, Matt Wiethoff, Dave Willner, Clemens Winter, Samuel Wolrich, Hannah Wong, Lauren Workman, Sherwin Wu, Jeff Wu, Michael Wu, Kai Xiao, Tao Xu, Sarah Yoo, Kevin Yu, Qiming Yuan, Wojciech Zaremba, Rowan Zellers, Chong Zhang, Marvin Zhang, Shengjia Zhao, Tianhao
  Zheng, Juntang Zhuang, William Zhuk, and Barret Zoph. 2023.
\newblock \href {http://arxiv.org/abs/2303.08774} {Gpt-4 technical report}.

\bibitem[{Ouyang et~al.(2022)Ouyang, Wu, Jiang, Almeida, Wainwright, Mishkin, Zhang, Agarwal, Slama, Ray, Schulman, Hilton, Kelton, Miller, Simens, Askell, Welinder, Christiano, Leike, and Lowe}]{instructgpt}
Long Ouyang, Jeff Wu, Xu~Jiang, Diogo Almeida, Carroll~L. Wainwright, Pamela Mishkin, Chong Zhang, Sandhini Agarwal, Katarina Slama, Alex Ray, John Schulman, Jacob Hilton, Fraser Kelton, Luke~E. Miller, Maddie Simens, Amanda Askell, Peter Welinder, Paul~Francis Christiano, Jan Leike, and Ryan~J. Lowe. 2022.
\newblock Training language models to follow instructions with human feedback.
\newblock \emph{ArXiv}, abs/2203.02155.

\bibitem[{Rozière et~al.(2023)Rozière, Gehring, Gloeckle, Sootla, Gat, Tan, Adi, Liu, Remez, Rapin, Kozhevnikov, Evtimov, Bitton, Bhatt, Ferrer, Grattafiori, Xiong, Défossez, Copet, Azhar, Touvron, Martin, Usunier, Scialom, and Synnaeve}]{codellama}
Baptiste Rozière, Jonas Gehring, Fabian Gloeckle, Sten Sootla, Itai Gat, Xiaoqing~Ellen Tan, Yossi Adi, Jingyu Liu, Tal Remez, Jérémy Rapin, Artyom Kozhevnikov, Ivan Evtimov, Joanna Bitton, Manish Bhatt, Cristian~Canton Ferrer, Aaron Grattafiori, Wenhan Xiong, Alexandre Défossez, Jade Copet, Faisal Azhar, Hugo Touvron, Louis Martin, Nicolas Usunier, Thomas Scialom, and Gabriel Synnaeve. 2023.
\newblock \href {http://arxiv.org/abs/2308.12950} {{Code Llama}: Open foundation models for code}.

\bibitem[{Shen et~al.(2023)Shen, Zhang, Chen, Zan, Geng, Fu, Zeng, Yu, Ji, Zhao, Guo, and Wang}]{pangucoder2}
Bo~Shen, Jiaxin Zhang, Taihong Chen, Daoguang Zan, Bing Geng, An~Fu, Muhan Zeng, Ailun Yu, Jichuan Ji, Jingyang Zhao, Yuenan Guo, and Qianxiang Wang. 2023.
\newblock \href {http://arxiv.org/abs/2307.14936} {{PanGu-Coder2}: Boosting large language models for code with ranking feedback}.

\bibitem[{Shinn et~al.(2023)Shinn, Cassano, Berman, Gopinath, Narasimhan, and Yao}]{shinn2023reflexion}
Noah Shinn, Federico Cassano, Edward Berman, Ashwin Gopinath, Karthik Narasimhan, and Shunyu Yao. 2023.
\newblock \href {http://arxiv.org/abs/2303.11366} {Reflexion: Language agents with verbal reinforcement learning}.

\bibitem[{Sweller(1988)}]{sweller1988cognitive}
John Sweller. 1988.
\newblock Cognitive load during problem solving: Effects on learning.
\newblock \emph{Cognitive science}, 12(2):257--285.

\bibitem[{Sweller(2011)}]{sweller2011cognitive}
John Sweller. 2011.
\newblock Cognitive load theory.
\newblock In \emph{Psychology of learning and motivation}, volume~55, pages 37--76. Elsevier.

\bibitem[{Turney(2000)}]{turney2000learning}
Peter~D Turney. 2000.
\newblock Learning algorithms for keyphrase extraction.
\newblock \emph{Information retrieval}, 2:303--336.

\bibitem[{Wan and Xiao(2008)}]{wan2008single}
Xiaojun Wan and Jianguo Xiao. 2008.
\newblock Single document keyphrase extraction using neighborhood knowledge.
\newblock In \emph{AAAI}, volume~8, pages 855--860.

\bibitem[{Wei et~al.(2023)Wei, Wang, Schuurmans, Bosma, Ichter, Xia, Chi, Le, and Zhou}]{cot}
Jason Wei, Xuezhi Wang, Dale Schuurmans, Maarten Bosma, Brian Ichter, Fei Xia, Ed~Chi, Quoc Le, and Denny Zhou. 2023.
\newblock \href {http://arxiv.org/abs/2201.11903} {Chain-of-thought prompting elicits reasoning in large language models}.

\bibitem[{Xu et~al.(2022)Xu, Alon, Neubig, and Hellendoorn}]{polycoder}
Frank~F. Xu, Uri Alon, Graham Neubig, and Vincent~J. Hellendoorn. 2022.
\newblock A systematic evaluation of large language models of code.
\newblock \emph{Proceedings of the 6th ACM SIGPLAN International Symposium on Machine Programming}.

\bibitem[{Zan et~al.(2022{\natexlab{a}})Zan, Chen, Lin, Guan, Wang, and Lou}]{apicoder}
Daoguang Zan, Bei Chen, Zeqi Lin, Bei Guan, Yongji Wang, and Jian-Guang Lou. 2022{\natexlab{a}}.
\newblock When language model meets private library.
\newblock In \emph{Conference on Empirical Methods in Natural Language Processing}.

\bibitem[{Zan et~al.(2022{\natexlab{b}})Zan, Chen, Yang, Lin, Kim, Guan, Wang, Chen, and Lou}]{cert}
Daoguang Zan, Bei Chen, Dejian Yang, Zeqi Lin, Minsu Kim, Bei Guan, Yongji Wang, Weizhu Chen, and Jian-Guang Lou. 2022{\natexlab{b}}.
\newblock {CERT}: Continual pre-training on sketches for library-oriented code generation.
\newblock In \emph{International Joint Conference on Artificial Intelligence}.

\bibitem[{Zan et~al.(2023)Zan, Chen, Zhang, Lu, Wu, Guan, Yongji, and Lou}]{nl2code}
Daoguang Zan, Bei Chen, Fengji Zhang, Dianjie Lu, Bingchao Wu, Bei Guan, Wang Yongji, and Jian-Guang Lou. 2023.
\newblock \href {https://aclanthology.org/2023.acl-long.411} {Large language models meet {NL}2{C}ode: A survey}.
\newblock In \emph{Proceedings of the 61st Annual Meeting of the Association for Computational Linguistics (Volume 1: Long Papers)}, pages 7443--7464, Toronto, Canada. Association for Computational Linguistics.

\bibitem[{Zhang et~al.(2023{\natexlab{a}})Zhang, Chen, Zhang, Liu, Zan, Mao, Lou, and Chen}]{repocoder}
Fengji Zhang, Bei Chen, Yue Zhang, Jin Liu, Daoguang Zan, Yi~Mao, Jian-Guang Lou, and Weizhu Chen. 2023{\natexlab{a}}.
\newblock \href {http://arxiv.org/abs/2303.12570} {{RepoCoder}: Repository-level code completion through iterative retrieval and generation}.

\bibitem[{Zhang et~al.(2023{\natexlab{b}})Zhang, Chen, Liu, Liao, Gong, Yu, Li, and Wang}]{zhang2023unifying}
Ziyin Zhang, Chaoyu Chen, Bingchang Liu, Cong Liao, Zi~Gong, Hang Yu, Jianguo Li, and Rui Wang. 2023{\natexlab{b}}.
\newblock \href {http://arxiv.org/abs/2311.07989} {Unifying the perspectives of nlp and software engineering: A survey on language models for code}.

\bibitem[{Zheng et~al.(2023{\natexlab{a}})Zheng, Xia, Zou, Dong, Wang, Xue, Wang, Shen, Wang, Li, Su, Yang, and Tang}]{codegeex}
Qinkai Zheng, Xiao Xia, Xu~Zou, Yuxiao Dong, Shanshan Wang, Yufei Xue, Zi-Yuan Wang, Lei Shen, Andi Wang, Yang Li, Teng Su, Zhilin Yang, and Jie Tang. 2023{\natexlab{a}}.
\newblock {CodeGeeX}: A pre-trained model for code generation with multilingual evaluations on humaneval-x.
\newblock \emph{ArXiv}, abs/2303.17568.

\bibitem[{Zheng et~al.(2023{\natexlab{b}})Zheng, Ning, Wang, Zhang, Zheng, Ye, and Chen}]{zheng2023survey}
Zibin Zheng, Kaiwen Ning, Yanlin Wang, Jingwen Zhang, Dewu Zheng, Mingxi Ye, and Jiachi Chen. 2023{\natexlab{b}}.
\newblock \href {http://arxiv.org/abs/2311.10372} {A survey of large language models for code: Evolution, benchmarking, and future trends}.

\end{thebibliography}
\bibliographystyle{acl_natbib}

\appendix
\section{Prompt Template}
\label{apx: main_result_template}

\subsection{\attentioncoder}
The prompt template of \attentioncoder can be divided into two types: one-chat and two-chat style. 
One-chat style is aimed to prompt LLMs by directly concatenating code context $\textbf{x}$ and extracted \attention $\mathcal{A}$ in one chat, 
while two-chat style means initially prompting the model to generate a solution for $\textbf{x}$ in the first chat round, and then prompting $\mathcal{A}$ in the subsequent chat round to refine the initial response.
These two types of templates are listed in Table~\ref{table: code_generation_templates}.

\begin{table*}[ht]
\small
\centering
\label{table: code_generation_templates}
\begin{tblr}{
  colspec = {Q[100]Q[350]Q[592]},
  cell{1}{1} = {c},
  cell{2}{1} = {c},
  cell{3}{1} = {c},
  cell{4}{1} = {c},
  cell{5}{1} = {c},
  cell{1}{2} = {c},
  cell{1}{3} = {c},
  hlines,
  vline{2-3} = {-}{},
}
\textbf{Type}           & \textbf{Prompt Template}                                                                                                                                                                                                                                                                                                                                                                                               & \textbf{ Example}                                                                                                                                                                                                                                                                                                                                                                                                                                                                                                                                                                                                                                                                                                                                                                                                                                                                                                                                                                                                                                                                                                                                                                                                                                                                 \\
One-chat & {Below is an instruction that describes a task. Write a response that appropriately completes the request.\\\#\#\# Instruction:\\Create a Python script for this problem:\\\textcolor[rgb]{0.502,0,0.502}{\{prompt with \textit{\textbf{Attention} putted into code comment}\}}\\\#\#\# Response:}                                                                                                                          & {Below is an instruction that describes a task, Write a response that appropriately completes the request.\\\#\#\# Instruction:\\Create a Python script for this problem:\\from typing import List\\\textcolor[rgb]{0.502,0,0.502}{def has\_close\_elements(numbers: List[float], threshold: float) - bool:~}\\\textcolor[rgb]{0.502,0,0.502}{~ ~ " Check if in given list of numbers, are any two numbers closer to each other than~}\\\textcolor[rgb]{0.502,0,0.502}{~ ~ given threshold.~}\\\textcolor[rgb]{0.502,0,0.502}{~ ~ Key Words:\textbf{ any two numbers, given list, given threshold~}}\\\textcolor[rgb]{0.502,0,0.502}{~ ~ has\_close\_elements([1.0, 2.0, 3.0], 0.5)~False~}\\\textcolor[rgb]{0.502,0,0.502}{~ ~ has\_close\_elements([1.0, 2.8, 3.0, 4.0, 5.0, 2.0], 0.3)~True~}\\\textcolor[rgb]{0.502,0,0.502}{~ ~ "}\\\#\#\# Response:}                                                                                                                                                                                                                                                                                                                                                                                              \\
Two-chat & {Question: \\\textcolor[rgb]{0.502,0,0.502}{\{origin prompt\}} \\Answer: \\\{generated code\} \\ Question: \\1. Check out if the target code generated before is correct to match the Code Description\\2. If not, pay attention to these Key Words in Code Description and rewrite the code to be correct\\\#\#\# Key Words\\\textcolor[rgb]{0.502,0,0.502}{\{\textit{Attention}\}}\\\#\#\# Code Description\\\textcolor[rgb]{0.502,0,0.502}{\{code comment\}}}                                                           & {Question:~\\\textcolor[rgb]{0.502,0,0.502}{from typing import List}\\\textcolor[rgb]{0.502,0,0.502}{def has\_close\_elements(numbers: List[float], threshold: float) - bool:~}\\\textcolor[rgb]{0.502,0,0.502}{~ ~ " Check if in given list of numbers, are any two numbers closer to each other than~}\\\textcolor[rgb]{0.502,0,0.502}{~ ~ given threshold.~}\\\textcolor[rgb]{0.502,0,0.502}{~ ~ Key Words: any two numbers, given list, given threshold~}\\\textcolor[rgb]{0.502,0,0.502}{~ ~ has\_close\_elements([1.0, 2.0, 3.0], 0.5)~False~}\\\textcolor[rgb]{0.502,0,0.502}{~ ~ has\_close\_elements([1.0, 2.8, 3.0, 4.0, 5.0, 2.0], 0.3)~True~}\\\textcolor[rgb]{0.502,0,0.502}{~ ~ "}\\Answer:~\\\{generated code\}\\Question:\\1. Check out if the target code generated before is correct to match the Code Description\\2. If not, pay attention to these Key Words in Code Description and rewrite the code to be correct\\\#\#\# Key Words\\\textcolor[rgb]{0.502,0,0.502}{any two numbers, given list, given threshold}\\\#\#\# Code Description\\\textcolor[rgb]{0.502,0,0.502}{Check if in given list of numbers, are any two numbers closer to each other than~given threshold.}\textcolor[rgb]{0.416,0.522,0.349}{}} 
\end{tblr}
\caption{Prompt Templates of \attentioncoder.}
\label{table: code_generation_templates}
\end{table*}

\subsection{LLMs Extraction}
\label{apx:llms extraction}
\begin{figure*}[t]
    \centering
    \includegraphics[width=0.75\linewidth]{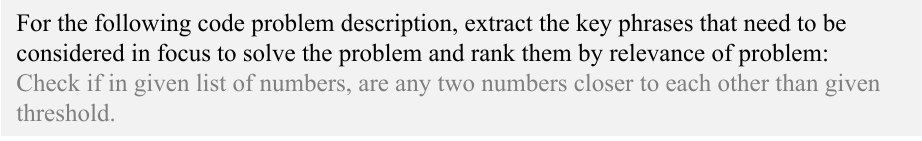}
    \caption{A specific prompt for extracting \attention using LLMs: code comment of context is in \textcolor{gray}{gray} background}
    \label{fig:gpt_extract_case}
\end{figure*}

We apply LLMs to extract key phrases of the code description to compare the extraction ability between LLMs and traditional NLP analysis tools. We conduct this experiment on GPT-3.5 with the prompt shown in Figure~\ref{fig:gpt_extract_case}. We post-process the generated result and acquire the extracted \attention.

\subsection{Three other tasks of transferability evaluation }
\label{apx:other_task_template}
\begin{figure*}[ht]
    \centering
    \includegraphics[width=0.75\linewidth]{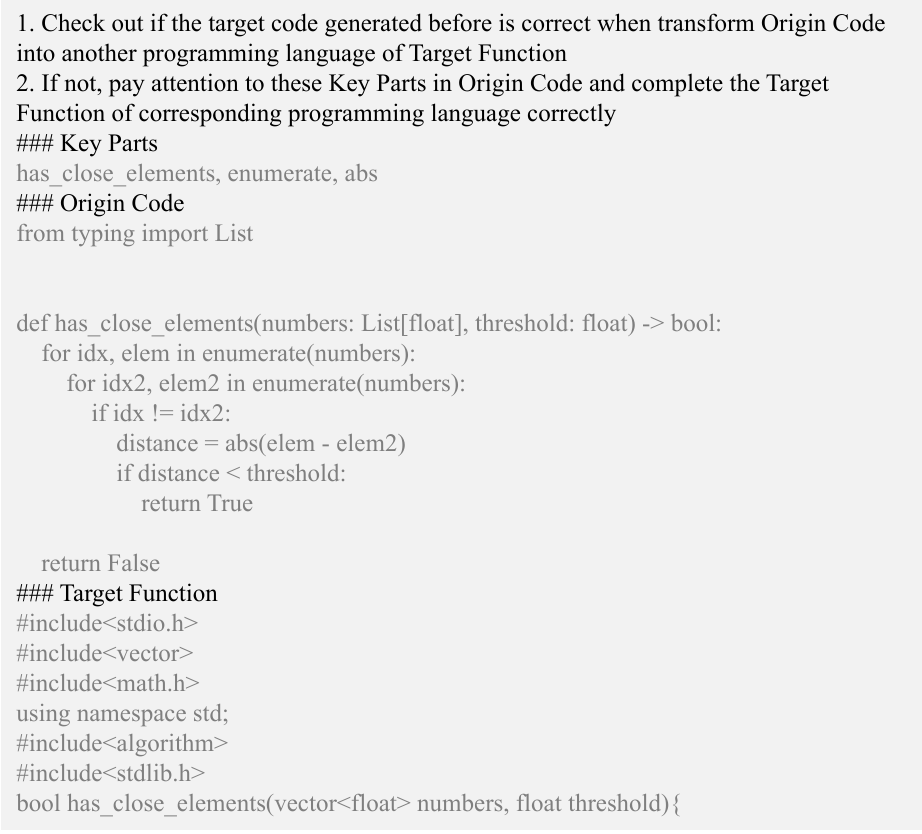}
    \caption{One example of code translation on python2c++.}
    \label{fig:code_translation}
\end{figure*}

\begin{figure*}[ht]
    \centering
    \includegraphics[width=0.75\linewidth]{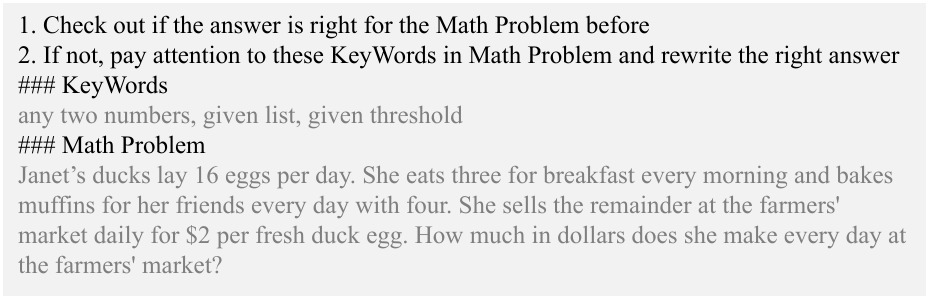}
    \caption{One concrete prompt of mathematical reasoning.}
    \label{fig:math_reasoning}
\end{figure*}

We apply our framework to other tasks including multiple-programming-language-oriented code generation, code translation, and mathematical reasoning.
The prompt template for the task of multiple-programming-language-oriented code generation is the same as the two-chat style template of the multiple-natural-language-oriented code generation task, as shown in Table~\ref{table: code_generation_templates}.
Prompt Templates of code translation and mathematical reasoning are shown in Figure~\ref{fig:code_translation} and Figure~\ref{fig:math_reasoning}, which are created by doing some task-related tweaks based on two-chat style prompt templates of code generation. It is worth noting that we extracted the function call as the \attention for the origin code in the code translation task since it doesn't have a natural language description.

\section{CoT and Reflexion Reproduction}
\label{apx:other_method_template}
\begin{figure*}[ht]
    \centering
    \includegraphics[width=0.75\linewidth]{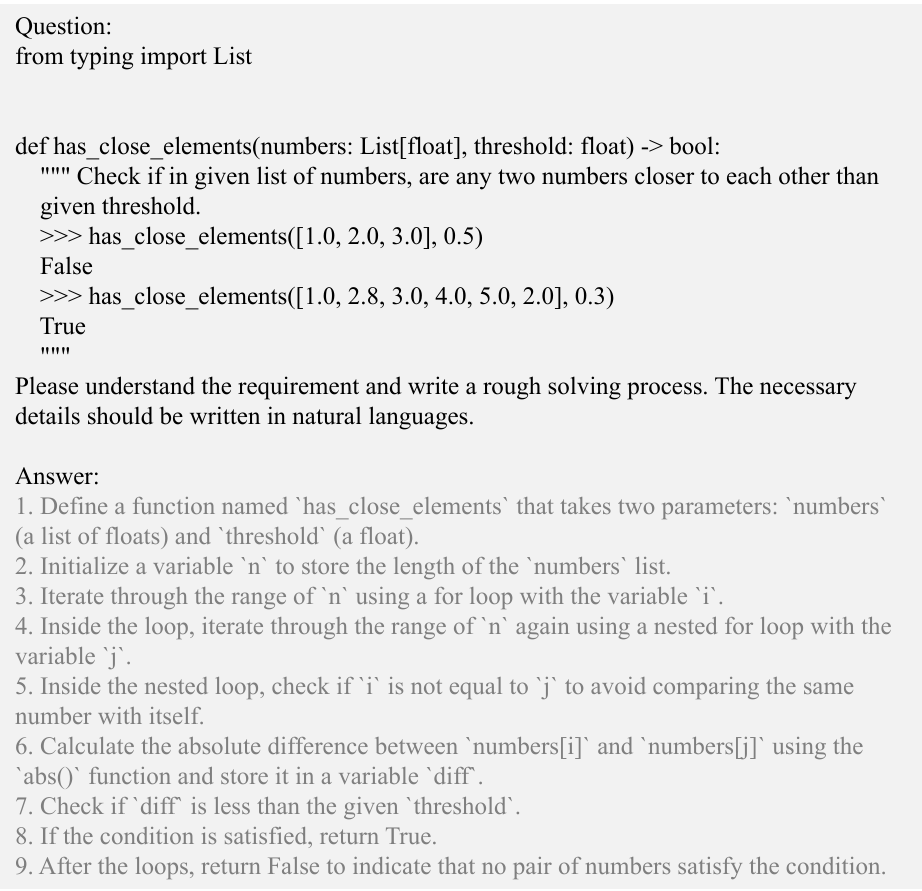}
    \caption{A case of generating a rough solving process in natural language: the solving process description is in \textcolor{gray}{gray} background.}
    \label{fig:generate_cot}
\end{figure*}

\begin{figure*}[ht]
    \centering
    \includegraphics[width=0.75\linewidth]{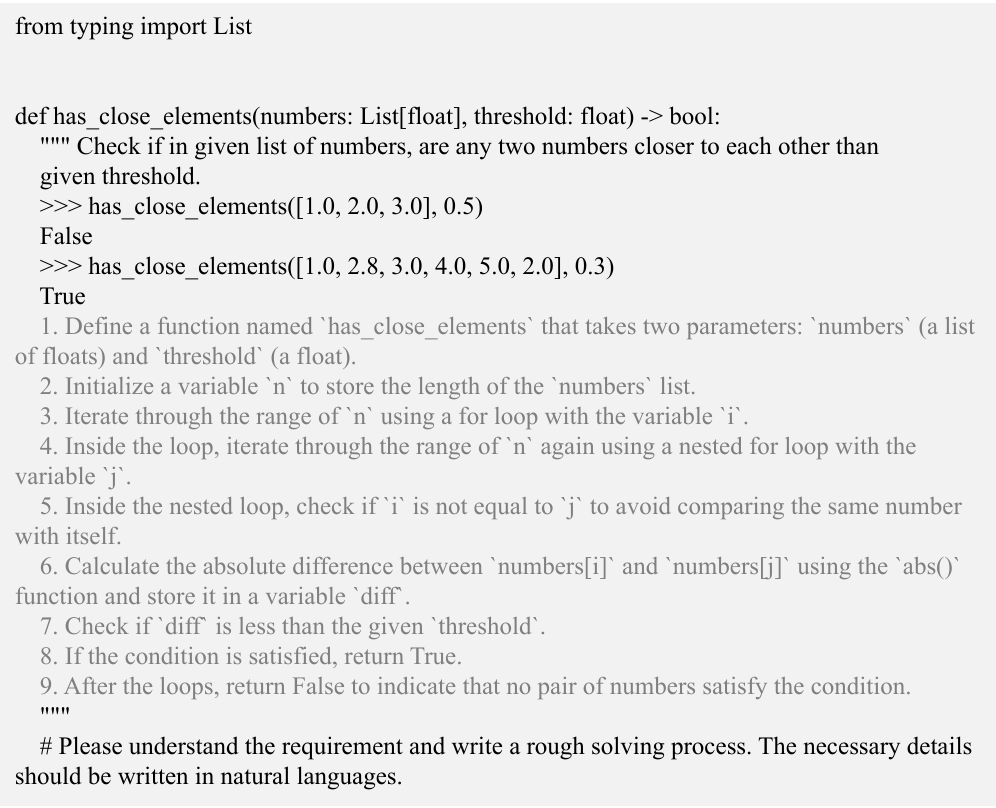}
    \caption{A case of generating target code using CoT method: the solving process description is in \textcolor{gray}{gray} background.}
    \label{fig:use_cot}
\end{figure*}

The Chain of Thought (CoT) method was initially proposed for mathematical reasoning and has subsequently been extended to encompass various prompt engineering tasks.
We reproduce CoT method exactly based on the implementation process of~\citet{scot}. A concrete case of generating and leveraging solving process procedures is respectively shown in Figure~\ref{fig:generate_cot} and~\ref{fig:use_cot}. 
We reproduce the Reflexion, using the open source project\footnote{\url{https://github.com/noahshinn/reflexion}} of \citet{shinn2023reflexion}. We follow the instructions of the project and do not change its source code. We only set the benchmark path to \multinlh to get the result of the framework running on multiple natural languages.

\begin{figure}[t]
    \centering
    \includegraphics[width=0.75\linewidth]{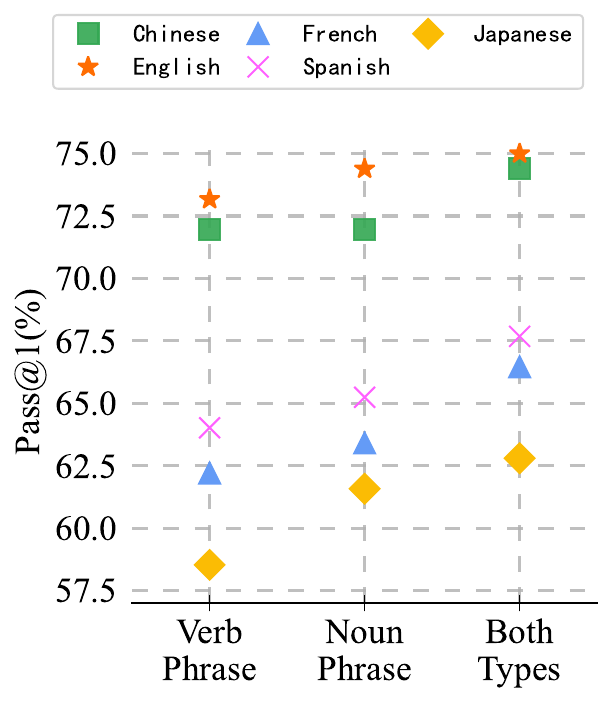}
    \caption{The result of Part of Speech of \attention: verb phrases, noun phrases, and both types.}
    \label{fig: attention type}
\end{figure}

\section{Part of Speech of \attention}
\label{apx: attention type}
The part of speech of \attention can be a noun or a verb.
We analyze the performance of these two parts of speech and the mixed setting (noun and verb) in Figure~\ref{fig: attention type}.
The results show that noun phrases are better than verb phrases.
After careful analysis, we find that the main reason is that 
noun phrases encompass more details that models are prone to err on and need to pay attention to.
For example, given ``Given a string s, count the number of uppercase vowels in even indices.'', \attentioncoder can extract verb phrases (``given a string s'', ``count the number'') and noun phrases (``uppercase vowels'', ``even indices'').
Moreover, from the results, we can find that the mixed setting is better than verb or noun since either verb or noun is meaningful for solving different problems.
Therefore, by default, this paper regards the mixed setting as \attention to prompt code LLMs.

\end{document}